\documentclass[letterpaper]{article} % DO NOT CHANGE THIS
\usepackage{aaai2027}
\usepackage[hyphens]{url}  % DO NOT CHANGE THIS
\usepackage{graphicx} % DO NOT CHANGE THIS
\usepackage{natbib}  % DO NOT CHANGE THIS AND DO NOT ADD ANY OPTIONS TO IT
\usepackage{caption} % DO NOT CHANGE THIS AND DO NOT ADD ANY OPTIONS TO IT
\usepackage{booktabs}

\usepackage{amsmath}
\usepackage{amssymb}

\usepackage{multirow}

\title{GAS-PFN: Graph-Aware Spatiotemporal Prior-Data Fitted Networks for Zero-Shot Traffic Forecasting}
\author{
    Joseph Natter,\textsuperscript{\rm 1}
    Long Vu,\textsuperscript{\rm 1}
    Yifan Zhang,\textsuperscript{\rm 2}
    Bobin Deng,\textsuperscript{\rm 1}
    Kun Suo,\textsuperscript{\rm 1}
    Rui Wu\textsuperscript{\rm 1}
}

\affiliations{
    \textsuperscript{\rm 1}Kennesaw State University\\
    \textsuperscript{\rm 2}Missouri State University
}

\begin{document}

\maketitle

\begin{abstract}

Traffic forecasting requires modeling both complex temporal dynamics and the spatial dependencies imposed by road topology. Supervised graph-based models achieve strong performance, but are typically trained for a particular dataset and sensor graph, making deployment to new or evolving networks costly. Prior-Data Fitted Networks (PFNs) offer a zero-shot alternative by learning in-context inference from synthetic task priors, which provide an explicit mechanism for encoding domain-specific assumptions. Existing forecasting PFNs, however, do not explicitly represent the graph-structured dependencies governing traffic dynamics.

To combine graph-aware traffic modeling with zero-shot adaptation, we introduce the \emph{Graph-Aware Spatiotemporal Prior-Data Fitted Network (GAS-PFN)}. GAS-PFN incorporates sampled road topology into its synthetic task prior through random-walk diffusion, teaching the model to learn graph-coupled dynamics during pretraining. Its co-designed inference architecture uses two mask-conditioned passes through a shared Transformer backbone to separately model sensor-specific temporal dynamics and spatiotemporal interactions among neighboring sensors before integrating their representations for prediction. This design enables GAS-PFN to use target-network structure and context without parameter updates.

Across six benchmarks at the 90-minute horizon, GAS-PFN reduces MSE by 34.0\% on average relative to TimePFN, its closest PFN-based predecessor. It also reduces MSE by 27.7\% on average relative to four recent industry-developed general-purpose zero-shot forecasters. In this comparison, all models receive 90 minutes of contiguous recent history, and any additional context is drawn from each dataset's chronological training split. These results indicate that aligning PFN priors and inference architectures with graph structure improves in-context learning for zero-shot traffic forecasting.
\end{abstract}


\section{Introduction}

Traffic conditions evolve over a road network, and congestion on one road can affect connected roads. Road topology shapes these interactions, but the patterns that actually emerge also depend on changing factors such as travel demand, incidents, weather, and human behavior, making them difficult to capture with a fully specified mechanistic model. Deep neural models provide a flexible means of learning these nonlinear spatiotemporal relationships directly from data, and graph neural networks and Transformer-based models have consequently achieved strong forecasting performance~\cite{li2018dcrnn,wu2020connecting,jiang2023pdformer,liu2023staeformer}.

Despite their strong performance, these models are typically trained for a particular dataset and sensor graph, requiring substantial data and computation when deployed to a new road network or adapted to changes in an existing one. Prior work reports that training a forecasting model on approximately one third of California's highway network required nearly eight GPU-days~\cite{mallick2020transfer,prabowo2023scpt}. Traffic-sensing deployments are also not static: sensors may be added, removed, or relocated, while sensor failures and communication interruptions can break the recent observation history even when older measurements remain available~\cite{zuo2023missing}. These conditions motivate forecasting methods that can adapt to a target network without repeated parameter optimization or reliance on a long uninterrupted recent history.

\begin{figure*}[!tb]
    \centering
    \includegraphics[width=\textwidth]{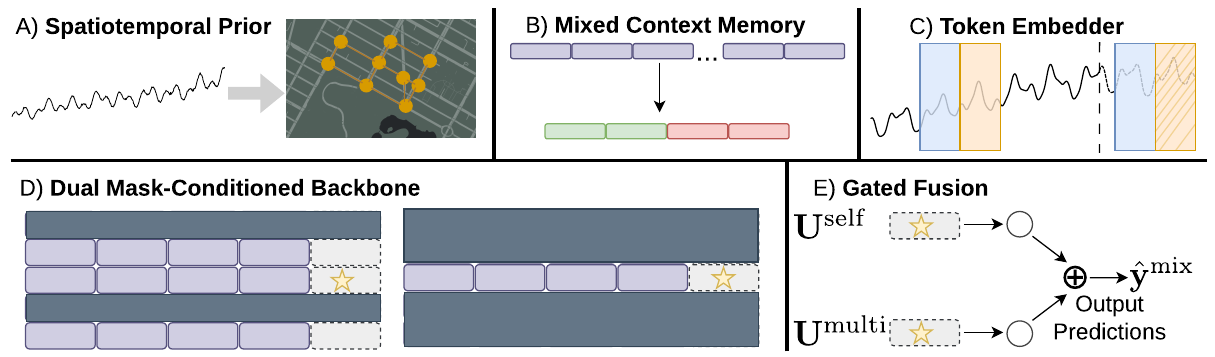}

    \caption{
    High-level overview of GAS-PFN. The method constructs graph-structured prior tasks, assembles
mixed-context memory, embeds context and query patches as tokens,
applies dual mask-conditioned backbone inference, and fuses the branch outputs to produce zero-shot forecasts.
    }
    \label{fig:pipeline}
\end{figure*}

Recent time-series foundation models, including Chronos-2, TimesFM-2.5, Moirai-2, and Lag-Llama, address the need for forecasting without task-specific retraining and have demonstrated strong zero-shot performance across heterogeneous datasets~\cite{ansari2025chronos2,das2024timesfm,liu2025moirai2, rasul2024lagllama}. However, these models are primarily designed as general-purpose sequence forecasters and do not explicitly incorporate the relational structure of road networks. This leaves a conceptual gap between transferable zero-shot forecasting and graph-aware traffic modeling.

Prior-Data Fitted Networks (PFNs) offer one emerging route toward bridging this gap by framing zero-shot prediction as in-context inference~\cite{muller2022transformers,hollmann2023tabpfn}. During pretraining, PFNs learn from synthetic tasks sampled from a prior to approximate Bayesian posterior predictive inference from context examples and a query. Their synthetic task priors provide an explicit mechanism for representing domain-specific assumptions, while inference-time conditioning enables adaptation without gradient-based optimization (i.e., training or fine-tuning). Recent work has extended this framework to time-series forecasting~\cite{taga2025timepfn}; however, existing time-series PFNs do not explicitly incorporate known graph topology into either their synthetic task priors or inference architectures.

Building on this idea, this work introduces the \emph{Graph-Aware Spatiotemporal Prior-Data Fitted Network (GAS-PFN)}, which proposes PFN-style forecasting for graph-structured traffic signals. GAS-PFN is pretrained on synthetic tasks in which interactions among channels are generated by diffusion over sampled road-network topologies. This exposes the model to spatially coupled temporal processes rather than treating each series as independent or only implicitly related. At inference time, GAS-PFN performs two differently masked passes through a shared Transformer backbone. A sensor-local pass restricts interactions to observations from the same sensor, capturing its intrinsic temporal dynamics. A neighborhood-aware pass allows interactions across both time and neighboring sensors, capturing graph-conditioned spatiotemporal dependencies. The resulting representations are then fused for prediction. By conditioning on both recent measurements and historical examples from the target network, GAS-PFN adapts through in-context learning rather than task-specific parameter updates.

Evaluation across six traffic forecasting benchmarks and three prediction horizons shows that GAS-PFN achieves the strongest overall zero-shot performance under the limited-recent-history protocol. At the 90-minute horizon, it obtains the best MAE and MSE on five of six datasets and achieves lower average MSE than both TimePFN and four recent general-purpose zero-shot forecasters. GAS-PFN also outperforms supervised SCPT when target-road training coverage is limited. These results indicate that encoding graph structure in both the synthetic prior and inference architecture improves zero-shot forecasting under limited recent context.

\begin{figure*}[!t]
    \centering
    \includegraphics[width=\textwidth,]{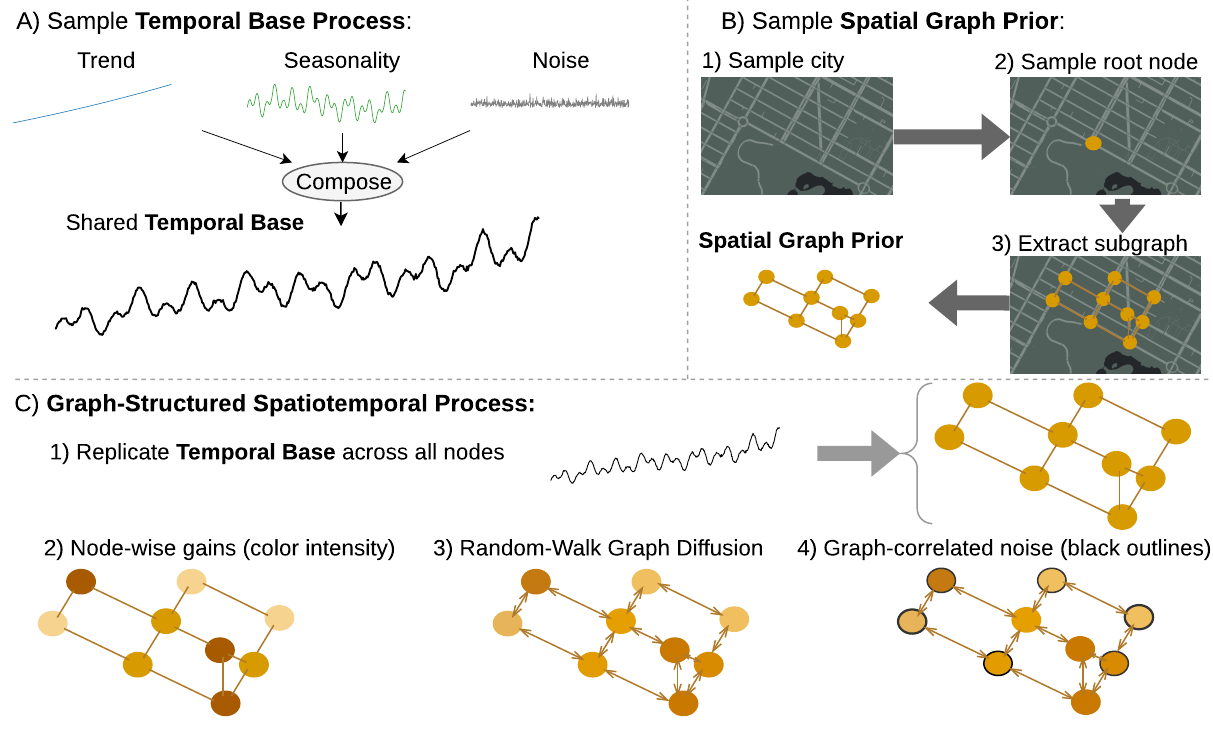}
    \caption{
    Overview of the proposed prior data generation process.
    (A) A shared temporal base is constructed by composing trend, seasonality, and noise components.
    (B) A spatial graph prior is sampled from a real-world road network by selecting a root node and extracting a connected subgraph.
    (C) The temporal base is lifted to all nodes, modulated by node-wise gains, propagated via a random-walk diffusion operator, and perturbed with graph-correlated noise to yield the final spatiotemporal process.
    }
    \label{fig:prior}
\end{figure*}

\textbf{Contributions.} The main contributions are as follows:
\begin{itemize}
    \item A graph-aware synthetic task prior that uses road-network topology and random-walk diffusion to expose a PFN to graph-coupled spatiotemporal dynamics during pretraining.
    
    \item A co-designed inference architecture that separately models sensor-specific temporal dynamics and spatiotemporal interactions among neighboring sensors, enabling the model to exploit its graph-aware prior during zero-shot prediction.

    \item A comprehensive evaluation across six traffic forecasting benchmarks and three prediction horizons, in which GAS-PFN achieves the best result in 27 of 36 MAE/MSE comparisons and outperforms supervised SCPT when target-road training coverage is limited.

\end{itemize}

\section{Related Work}

\subsection{Zero-Shot Time-Series Forecasting}

Recent time-series foundation models have demonstrated strong zero-shot forecasting performance across heterogeneous datasets. Chronos-2 uses group attention to support univariate, multivariate, and covariate-informed forecasting within a single pretrained model~\cite{ansari2025chronos2}. TimesFM~2.5 builds on a patch-based decoder-only Transformer trained for zero-shot forecasting across domains, horizons, and temporal resolutions~\cite{das2024timesfm}. Moirai~2.0 adopts a decoder-only architecture with quantile forecasting and multi-token prediction to improve both forecasting accuracy and inference efficiency~\cite{liu2025moirai2}. These models support general-purpose zero-shot forecasting but do not
explicitly incorporate known relational structure between channels,
such as road-network topology. GAS-PFN instead investigates how such domain structure can be encoded directly into zero-shot pretraining and inference.

\subsection{Traffic Forecasting and Cross-Network Generalization}

Traffic forecasting methods commonly exploit spatial dependencies among sensors through graph-based architectures. DCRNN models traffic propagation as diffusion over a directed graph and combines graph convolution with recurrent temporal modeling~\cite{li2018dcrnn}. Graph WaveNet introduces adaptive adjacency learning and dilated temporal convolutions to capture spatial relationships and long-range temporal dependencies~\cite{wu2019graphwavenet}. More recent Transformer-based methods include PDFormer, which incorporates propagation-delay-aware spatial attention~\cite{jiang2023pdformer}, and STAEformer, which obtains strong performance through learned spatiotemporal embeddings~\cite{liu2023staeformer}.

These models achieve strong supervised performance but are generally optimized for a particular dataset and sensor graph. Graph-dependent components such as node embeddings, adjacency representations, and attention masks may therefore be tied to the network observed during training, limiting direct application to altered node sets or unseen road networks. Transfer-oriented methods address this limitation by learning representations intended to generalize beyond the roads observed during training. Mallick et al.\ transfer a graph neural traffic forecaster between highway networks~\cite{mallick2020transfer}, while SCPT contrastively pretrains a spatial encoder to infer representations for roads excluded from supervised training~\cite{prabowo2023scpt}. These methods learn representations that generalize to new roads, whereas GAS-PFN adapts at inference time through context examples from the target graph.

\subsection{Prior-Data Fitted Networks}

Prior-Data Fitted Networks (PFNs) learn to approximate Bayesian posterior predictive inference by training on synthetic tasks sampled from a predefined prior~\cite{muller2022transformers,hollmann2023tabpfn}. At inference time, predictions are produced by conditioning on observed context examples rather than updating model parameters. ForecastPFN applied this approach to zero-shot forecasting using a fully synthetic time-series distribution~\cite{dooley2023forecastpfn}. LaT-PFN combines PFN-based inference with a joint-embedding predictive architecture, performing in-context learning within a prediction-oriented latent space~\cite{verdenius2024latpfn}. TimePFN extends PFN forecasting to multivariate series through synthetic Gaussian-process tasks and an architecture that models temporal and cross-channel dependencies~\cite{taga2025timepfn}, making it the closest PFN-based predecessor to GAS-PFN.

A distinguishing feature of PFNs is that the synthetic task prior constitutes an explicit modeling choice through which domain-specific assumptions can be introduced. Existing forecasting PFNs, however, do not encode graph topology within their synthetic task priors and inference architectures. GAS-PFN addresses this gap by co-designing graph-structured synthetic tasks and graph-aware in-context inference for traffic forecasting.

\section{Method}
\label{sec:method}

GAS-PFN combines a graph-aware synthetic task prior with an inference architecture designed to exploit graph-structured context. Figure~\ref{fig:pipeline} summarizes the full pipeline. We first formalize the zero-shot PFN setting, then introduce a graph-aware spatiotemporal prior that exposes the model to traffic-like temporal dynamics and graph-coupled interactions, followed by an architecture designed to learn these patterns during pretraining and infer them from context at test time.

\subsection{Problem Setting and PFN Inference}
\label{sec:setting}

Consider forecasting over a graph $G=(V,E)$, where each node $v\in V$ corresponds to a sensor and $\mathbf{y}(t)\in\mathbb{R}^{|V|}$ denotes the graph signal observed at time $t$. For a node $v$ and temporal endpoint $\tau$, we define a lookback patch and prediction target as
\begin{equation}
\begin{aligned}
\mathbf{x}_{v,\tau}
&=
[y_v(\tau-L+1),\ldots,y_v(\tau)]
\in\mathbb{R}^{L},\\
\mathbf{y}_{v,\tau}
&=
[y_v(\tau+1),\ldots,y_v(\tau+H)]
\in\mathbb{R}^{H},
\end{aligned}
\label{eq:patches}
\end{equation}
where $L$ and $H$ are the lookback and forecast lengths.

For each node $v\in V$, let $\mathcal{C}_v$ denote the set of
context-window endpoints selected for that node. The context set is
\begin{equation}
\mathcal{D}_{\mathrm{ctx}}
=
\left\{
(\mathbf{x}_{v,\tau},\mathbf{y}_{v,\tau})
:
v\in V,\;
\tau\in\mathcal{C}_v
\right\}.
\label{eq:context-set}
\end{equation}

At inference time, the model conditions on $\mathcal{D}_{\mathrm{ctx}}$
and a query lookback patch $\mathbf{x}_{v,\tau}$ whose target
$\mathbf{y}_{v,\tau}$ is unobserved. The resulting prediction is
\begin{equation}
\hat{\mathbf{y}}_{v,\tau}
=
q_{\theta}
\left(
\mathbf{x}_{v,\tau}
\mid
\mathcal{D}_{\mathrm{ctx}},G
\right),
\end{equation}
where $\theta$ remains fixed when the model is applied to a target dataset. In this work, \emph{zero-shot} denotes inference without target-dataset
parameter updates; context examples may be drawn from the target
dataset's chronological training split.

Following the PFN framework~\cite{muller2022transformers}, $q_{\theta}$
is pretrained on synthetic tasks sampled from a prior $\Pi$ and learns
to approximate the posterior predictive induced by $\Pi$ through
context rather than task-specific parameter updates. The node-local patch representation keeps token dimensionality independent of $|V|$, allowing the same model to process sensor graphs of different sizes.

\begin{figure}[!t]
    \centering
    \includegraphics[
        width=\columnwidth,
    ]{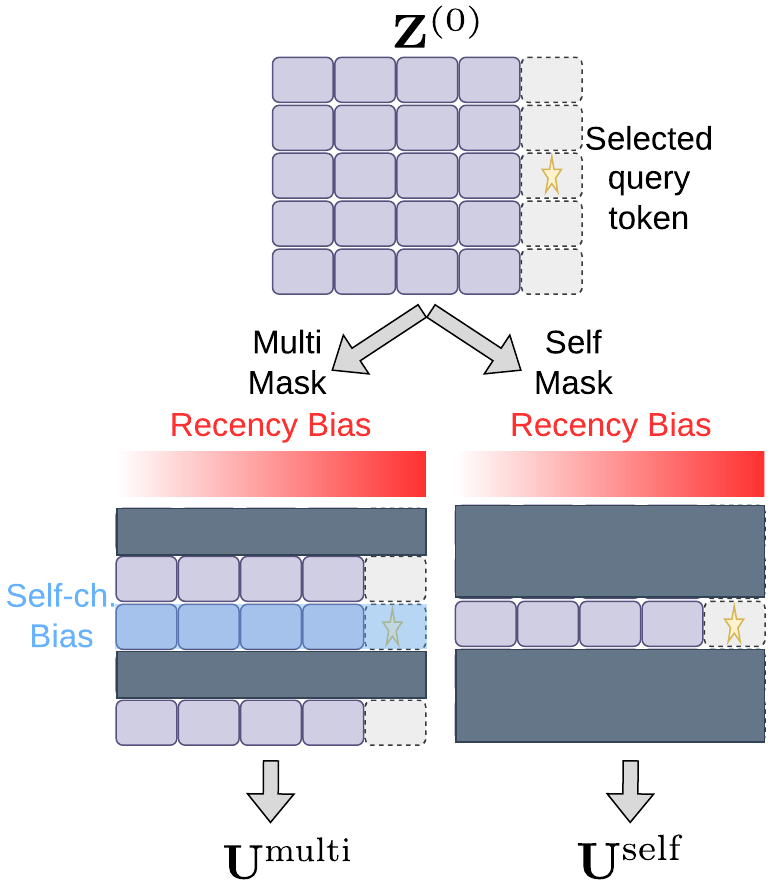}
\caption{
Dual mask-conditioned backbone. A shared Transformer backbone evaluates
$\mathbf{Z}^{(0)}$ under separate multi-channel and self-channel attention masks. Each branch shows the mask and attention biases for the selected query token; gray blocks denote masked keys. Only the resulting query-token representations, $\mathbf{U}^{\mathrm{multi}}$ and $\mathbf{U}^{\mathrm{self}}$, are retained.
}
    \label{fig:backbone}
\end{figure}

\subsection{Spatiotemporal Prior}
\label{sec:prior}

The central prior-design principle is to expose the PFN to temporal processes whose channels interact over realistic road-network structures. Each synthetic task combines a compositional temporal process with a connected graph sampled from a bank of real road networks, as illustrated in figure~\ref{fig:prior}.

The univariate temporal generator adapts the compositional prior used by
LaT-PFN, itself derived from ForecastPFN~\cite{verdenius2024latpfn,
dooley2023forecastpfn}. We modify its composition rule, parameter
distributions, and exponential-trend construction, and extend it with
road-network sampling and graph-structured spatial perturbations to
produce multivariate traffic processes.

A univariate temporal base $y_0(t)$ is formed by composing trend,
seasonal, and stochastic components sampled from broadly defined
distributions. The sampled graph $G$ is obtained by selecting a road network, choosing a root node, and extracting a connected subgraph. Detailed component distributions and graph-sampling procedures are provided in Appendix~A.

Let $\mathbf{A}$ denote the adjacency matrix of $G$, and define the random-walk transition matrix
\begin{equation}
\mathbf{P}
=
\mathbf{D}^{-1}(\mathbf{A}+\mathbf{I}),
\end{equation}
where $\mathbf{D}$ is the corresponding degree matrix. We use the lazy random-walk diffusion operator
\begin{equation}
\mathcal{P}(\mathbf{u})
=
\frac{1}{2}
\left(
\mathbf{u}+\mathbf{P}\mathbf{u}
\right),
\end{equation}
which locally smooths a graph signal while retaining part of its original value~\cite{shuman2013emerging,ortega2018graph}. This retains an explicit identity contribution while applying one step
of neighborhood averaging, limiting potential over-smoothing.

The temporal base is replicated across nodes as
\begin{equation}
\mathbf{y}_0(t)=y_0(t)\mathbf{1}.
\end{equation}
Node-specific gains $\mathbf{g}$ introduce local variation,
\begin{equation}
g_v \sim 1+\sigma_g\mathcal{N}(0,1),
\qquad
\mathbf{z}(t)=\mathbf{g}\odot\mathbf{y}_0(t),
\end{equation}
after which graph propagation produces
\begin{equation}
\mathbf{z}_{\mathrm{prop}}(t)
=
(1-\beta)\mathbf{z}(t)
+
\beta\mathcal{P}(\mathbf{z}(t)).
\end{equation}
We additionally sample Gaussian noise
$\mathbf{n}(t)\sim\mathcal{N}(\mathbf{0},\mathbf{I})$
and apply a graph-filtering operator:
\begin{equation}
\boldsymbol{\epsilon}_{G}(t)
=
\mathcal{P}_{\epsilon}(\mathbf{n}(t)),
\end{equation}
where the noise-filtering operator
$\mathcal{P}_{\epsilon}$ is
defined in Appendix~A.
The final graph signal is
\begin{equation}
\mathbf{y}(t)
=
\mathbf{y}_0(t)
+
\alpha
\left[
\mathbf{z}_{\mathrm{prop}}(t)
-
\mathbf{y}_0(t)
+
\boldsymbol{\epsilon}_{G}(t)
\right],
\label{eq:prior-final}
\end{equation}
where $\alpha$ controls the overall strength of spatial variation and $\beta$ controls graph propagation.

Each sampled task therefore combines shared temporal behavior with node-specific and graph-correlated deviations. By biasing the synthetic task distribution toward graph-coupled functions, this construction is intended to reduce prior mismatch when target traffic dynamics exhibit local road-network dependence, encouraging the PFN to infer both temporal dynamics and spatial interactions from context.

\subsection{Graph-Aware ICL Architecture}
\label{sec:architecture}

\paragraph{Mixed-context memory and token embedding.}
For each node, GAS-PFN constructs a mixed-context memory from two equally
sized sources of historical context: the $K$ most recent windows from the dataset's
chronological training split and $K$ older anchor windows distributed across the
remaining training split. Let
$\mathcal{C}^{\mathrm{recent}}$
and
$\mathcal{C}^{\mathrm{anchor}}_v$
denote their respective window endpoints. The per-node context set is
\begin{equation}
\widetilde{\mathcal{C}}_v
=
\mathcal{C}^{\mathrm{recent}}
\cup
\mathcal{C}^{\mathrm{anchor}}_v,
\qquad
\left|\mathcal{C}^{\mathrm{recent}}\right|
=
\left|\mathcal{C}^{\mathrm{anchor}}_v\right|
=
K.
\label{eq:mixed-context}
\end{equation}
Recent windows use common timestamps across sensors, whereas historical anchors are sampled across chronological strata independently for each node and need not be time-aligned. We use $K=16$, yielding 32 historical support patches per node.

Let $\mathcal{Q}$ denote the set of query-window endpoints.
The selected context patches and query windows are embedded as
\begin{equation}
\begin{aligned}
\mathbf{e}^{\mathrm{ctx}}_{v,\tau}
&=
\phi_{\mathrm{ctx}}
\left(
[\mathbf{x}_{v,\tau};\mathbf{y}_{v,\tau}]
\right),
&&
\tau\in\widetilde{\mathcal{C}}_v,
\\
\mathbf{e}^{\mathrm{qry}}_{v,\tau}
&=
\phi_{\mathrm{qry}}
\left(
[\mathbf{x}_{v,\tau};\mathbf{0}_{H}]
\right),
&&
\tau\in\mathcal{Q}.
\end{aligned}
\label{eq:tokenization}
\end{equation}
where $\phi_{\mathrm{ctx}}$ and $\phi_{\mathrm{qry}}$ are learned affine
projections. Each token retains its node index and window-end timestamp
for attention masking.

Let $\mathbf{E}^{\mathrm{ctx}}$ denote the embeddings of all selected
context tokens and $\mathbf{E}^{\mathrm{qry}}$ the embeddings of the
query tokens. The Transformer input is
\begin{equation}
\mathbf{Z}^{(0)}
=
\left[
\mathbf{E}^{\mathrm{ctx}};
\mathbf{E}^{\mathrm{qry}}
\right].
\label{eq:transformer-input}
\end{equation}

\begin{table*}[t]
\centering
\footnotesize
\setlength{\tabcolsep}{2.2pt}
\renewcommand{\arraystretch}{1.08}

\begin{tabular}{cc*{6}{cc}}
\toprule
\multirow{2}{*}{\textbf{Dataset}} &
\multirow{2}{*}{\textbf{Horizon}} &
\multicolumn{2}{c}{\textbf{GAS-PFN}} &
\multicolumn{2}{c}{\textbf{Chronos-2}} &
\multicolumn{2}{c}{\textbf{TimesFM-2.5}} &
\multicolumn{2}{c}{\textbf{Moirai-2}} &
\multicolumn{2}{c}{\textbf{Lag-Llama}} &
\multicolumn{2}{c}{\textbf{TimePFN}} \\
\cmidrule(lr){3-4}
\cmidrule(lr){5-6}
\cmidrule(lr){7-8}
\cmidrule(lr){9-10}
\cmidrule(lr){11-12}
\cmidrule(lr){13-14}
& &
\textbf{MAE} & \textbf{MSE} &
\textbf{MAE} & \textbf{MSE} &
\textbf{MAE} & \textbf{MSE} &
\textbf{MAE} & \textbf{MSE} &
\textbf{MAE} & \textbf{MSE} &
\textbf{MAE} & \textbf{MSE} \\
\midrule

\multirow{3}{*}{METR-LA}
& 30 min
& 3.97$\pm$0.05 & \textbf{68.31}$\pm$1.92
& \textbf{3.68} & 70.62
& \underline{3.80} & 72.30
& 4.23 & 83.92
& 4.86 & 86.23
& 4.86 & \underline{69.41} \\

& 60 min
& 4.83$\pm$0.06 & 97.61$\pm$2.92
& \textbf{4.38} & \textbf{91.23}
& \underline{4.73} & 108.38
& 5.30 & 124.56
& 6.21 & 130.31
& 5.59 & \underline{94.33} \\

& 90 min
& 5.50$\pm$0.05 & 119.69$\pm$4.15
& \textbf{4.95} & \textbf{107.42}
& \underline{5.45} & 136.04
& 6.35 & 163.35
& 7.41 & 178.32
& 6.13 & \underline{112.73} \\

\midrule

\multirow{3}{*}{PEMS-BAY}
& 30 min
& 1.73$\pm$0.01 & \underline{14.85}$\pm$0.09
& \textbf{1.61} & \textbf{12.90}
& \underline{1.65} & 15.01
& 1.74 & 15.53
& 2.60 & 45.83
& 2.41 & 22.35 \\

& 60 min
& \underline{2.21}$\pm$0.01 & \textbf{25.22}$\pm$0.09
& 2.24 & \underline{27.25}
& \textbf{2.18} & 27.70
& 2.36 & 29.99
& 2.99 & 52.64
& 2.85 & 32.51 \\

& 90 min
& \textbf{2.59}$\pm$0.01 & \textbf{34.45}$\pm$0.08
& 2.71 & 40.58
& \underline{2.62} & \underline{39.32}
& 2.91 & 44.67
& 3.29 & 58.78
& 3.21 & 41.51 \\

\midrule

\multirow{3}{*}{PEMS03}
& 30 min
& \textbf{17.57}$\pm$0.49 & \underline{749.58}$\pm$29.82
& \underline{17.74} & \textbf{747.80}
& 19.52 & 894.14
& 19.64 & 913.24
& 20.23 & 918.27
& 34.50 & 2201.97 \\

& 60 min
& \textbf{21.51}$\pm$0.75 & \textbf{1160.61}$\pm$56.57
& \underline{23.22} & \underline{1320.70}
& 25.26 & 1496.84
& 25.74 & 1646.29
& 27.25 & 1672.31
& 38.46 & 2788.31 \\

& 90 min
& \textbf{25.97}$\pm$1.10 & \textbf{1740.93}$\pm$105.99
& \underline{28.98} & \underline{2080.14}
& 31.61 & 2357.27
& 32.36 & 2666.41
& 34.66 & 2709.34
& 42.60 & 3417.89 \\

\midrule

\multirow{3}{*}{PEMS04}
& 30 min
& \textbf{22.52}$\pm$0.23 & \textbf{1243.43}$\pm$10.90
& \underline{24.24} & \underline{1402.54}
& 25.25 & 1586.96
& 25.90 & 1518.06
& 27.62 & 1698.79
& 43.19 & 3218.50 \\

& 60 min
& \textbf{27.16}$\pm$0.34 & \textbf{1782.98}$\pm$21.60
& \underline{31.26} & \underline{2300.66}
& 31.77 & 2514.64
& 33.45 & 2523.53
& 38.54 & 3152.43
& 47.91 & 3987.86 \\

& 90 min
& \textbf{32.32}$\pm$0.43 & \textbf{2517.88}$\pm$44.66
& \underline{37.92} & \underline{3370.61}
& 38.58 & 3708.20
& 41.73 & 3966.89
& 50.72 & 5371.03
& 52.92 & 4841.89 \\

\midrule

\multirow{3}{*}{PEMS07}
& 30 min
& \textbf{25.15}$\pm$0.44 & \textbf{1493.64}$\pm$27.66
& \underline{25.95} & \underline{1522.69}
& 26.25 & 1612.32
& 28.74 & 1858.14
& 29.20 & 1911.20
& 46.42 & 4035.22 \\

& 60 min
& \textbf{30.81}$\pm$0.61 & \textbf{2223.93}$\pm$45.83
& \underline{32.76} & \underline{2366.80}
& 33.96 & 2627.48
& 38.36 & 3342.03
& 39.99 & 3328.94
& 52.46 & 5170.30 \\

& 90 min
& \textbf{37.15}$\pm$0.80 & \textbf{3239.16}$\pm$77.26
& \underline{40.01} & \underline{3558.66}
& 42.17 & 4053.21
& 48.91 & 5490.79
& 50.77 & 5272.50
& 58.75 & 6523.84 \\

\midrule

\multirow{3}{*}{PEMS08}
& 30 min
& \textbf{17.78}$\pm$0.15 & \textbf{767.87}$\pm$14.13
& 19.10 & 834.07
& \underline{18.50} & \underline{816.30}
& 20.44 & 961.84
& 22.67 & 1121.92
& 26.94 & 1488.01 \\

& 60 min
& \textbf{21.35}$\pm$0.30 & \textbf{1112.37}$\pm$33.62
& 24.15 & \underline{1305.08}
& \underline{23.54} & 1313.15
& 26.77 & 1666.28
& 32.68 & 2297.85
& 31.22 & 2031.89 \\

& 90 min
& \textbf{25.21}$\pm$0.38 & \textbf{1548.52}$\pm$58.23
& 29.44 & \underline{1922.89}
& \underline{28.60} & 1924.25
& 33.96 & 2695.22
& 43.51 & 4055.36
& 36.54 & 2864.76 \\

\midrule

\multicolumn{2}{l}{\textbf{Overall metric wins}}
& \multicolumn{2}{c}{\textbf{27}}
& \multicolumn{2}{c}{\underline{8}}
& \multicolumn{2}{c}{1}
& \multicolumn{2}{c}{0}
& \multicolumn{2}{c}{0}
& \multicolumn{2}{c}{0} \\

\bottomrule
\end{tabular}

\vspace{2pt}

\caption{Zero-shot forecasting performance across datasets and prediction horizons. Lower values indicate better performance. Best results are shown in bold and second-best results are underlined. GAS-PFN reports mean $\pm$ standard deviation over three independently trained checkpoints; each baseline uses a single released pretrained checkpoint.}
\label{tab:horizon_results}
\end{table*}

\paragraph{Dual mask-conditioned backbone.}
GAS-PFN evaluates the same Transformer backbone under two attention masks:
\begin{equation}
\begin{aligned}
\mathbf{U}^{\mathrm{self}}
&=
f_{\theta}
(\mathbf{Z}^{(0)};\mathcal{M}^{\mathrm{self}})_{\mathrm{qry}},\\
\mathbf{U}^{\mathrm{multi}}
&=
f_{\theta}
(\mathbf{Z}^{(0)};\mathcal{M}^{\mathrm{multi}})_{\mathrm{qry}}.
\end{aligned}
\label{eq:dual-backbone}
\end{equation}
The two passes share backbone parameters but impose different structural constraints, as shown in figure~\ref{fig:backbone}.

The self mask restricts attention to causally valid tokens from the same
sensor and disables query--query attention, thereby modeling
sensor-local temporal dynamics. The multi-channel mask permits attention
within a selected sensor neighborhood and, in the reported evaluation
protocol, allows query--query interactions only among tokens at the same
forecast time. Historical anchors retain their original window
timestamps, preserving causal ordering even when anchors are not aligned
across sensors.

For token pairs permitted by the corresponding mask, the pre-softmax
attention logit for a given head and branch is
\begin{equation}
\ell_{ij}
=
\frac{\mathbf{q}_{i}^{\top}\mathbf{k}_{j}}{\sqrt{d_h}}
-
\gamma\max(t_i-t_j,0)
+
b_{\mathrm{same}}\,\delta(ch_i,ch_j),
\label{eq:attention-logit}
\end{equation}
where $\delta(ch_i,ch_j)=1$ when tokens $i$ and $j$ belong to the same sensor and $0$ otherwise. Each head learns its own temporal-distance coefficient $\gamma$ and self-channel bias $b_{\mathrm{same}}$. In the self pass, the self-channel term is constant across all visible keys and therefore cancels under the softmax; in the multi-channel pass, it distinguishes self-channel from neighboring-channel interactions.

Neighborhoods combine known graph connectivity with semantic similarity. Let $\overline{\mathbf{A}}$ denote the
row-scaled task-local graph-score matrix and
$\mathbf{S}^{\mathrm{sem}}$ the cosine-similarity matrix computed from
encoded, aggregated channel representations. We define
\begin{equation}
\mathbf{S}^{\mathrm{hyb}}
=
(1-\eta)\overline{\mathbf{A}}
+
\eta\mathbf{S}^{\mathrm{sem}},
\qquad
\mathcal{N}_v
=
\operatorname{TopK}
(\mathbf{S}^{\mathrm{hyb}}_{v,:}),
\label{eq:hybrid-neighborhood}
\end{equation}
where $\eta$ balances the graph-connectivity and semantic-similarity
terms. The selected channel neighborhoods are converted into token-level
attention constraints for the multi-channel mask. Further details are
provided in Appendix~B.

\paragraph{Fusion and training objective.}
A learned gate combines the two query representations:
\begin{equation}
\begin{aligned}
\mathbf{g}
&=
\sigma
\left(
g_{\theta}
[
\mathbf{U}^{\mathrm{self}};
\mathbf{U}^{\mathrm{multi}}
]
\right),\\
\mathbf{U}^{\mathrm{mix}}
&=
(1-\mathbf{g})\odot\mathbf{U}^{\mathrm{self}}
+
\mathbf{g}\odot\mathbf{U}^{\mathrm{multi}},\\
\widehat{\mathbf{Y}}^{\mathrm{mix}}
&=
h_{\theta}(\mathbf{U}^{\mathrm{mix}}),
\end{aligned}
\label{eq:fusion}
\end{equation}
where $h_{\theta}$ is a shared output projection.

The same projection produces auxiliary predictions
$\widehat{\mathbf{Y}}^{\mathrm{self}}$
and
$\widehat{\mathbf{Y}}^{\mathrm{multi}}$
from the individual branch representations.

Training minimizes
\begin{equation}
\begin{aligned}
\mathcal{L}
={}&
\mathrm{MSE}
\left(
\widehat{\mathbf{Y}}^{\mathrm{mix}},
\mathbf{Y}
\right)
\\
&+
\lambda_{\mathrm{aux}}
\mathrm{MSE}
\left(
\widehat{\mathbf{Y}}^{\mathrm{self}},
\mathbf{Y}
\right)
\\
&+
\lambda_{\mathrm{aux}}
\mathrm{MSE}
\left(
\widehat{\mathbf{Y}}^{\mathrm{multi}},
\mathbf{Y}
\right)
\\
&-
\lambda_{\mathrm{ent}}
\mathcal{H}(\mathbf{g}).
\end{aligned}
\label{eq:objective}
\end{equation}
where the auxiliary losses keep both branches individually predictive and the entropy term discourages premature gate collapse. At inference time, only the fused prediction is returned.

\section{Results}

\paragraph{Datasets and metrics.}
We evaluate GAS-PFN on six widely used traffic forecasting benchmarks, all sampled at five-minute intervals. METR-LA contains traffic-speed measurements from 207 highway sensors in Los Angeles County collected from March through June 2012, while PEMS-BAY contains speed measurements from 325 sensors in the San Francisco Bay Area collected from January through May 2017~\cite{li2018dcrnn}. PEMS03, PEMS04, PEMS07, and PEMS08 contain traffic-flow measurements from 358, 307, 883, and 170 sensors, respectively, drawn from the corresponding California Performance Measurement System districts~\cite{chen2001pems}. Their collection periods span September--November 2018, January--February 2018, May--August 2017, and July--August 2016, respectively~\cite{guo2019astgcn}. These benchmarks provide a standard basis for comparison
while covering both speed and flow prediction across road networks of different sizes and geographic settings. We use chronological 70\%/10\%/20\% train, validation, and test splits and report MAE and MSE cumulatively at horizons of 6, 12, and 18 steps, corresponding to 30, 60,
and 90 minutes;
MAE summarizes typical absolute error, whereas MSE places greater weight
on large deviations. Following standard traffic-forecasting practice, zero-valued targets are masked when computing the training loss and evaluation metrics, as zero commonly serves as the missing-value marker in these benchmarks~\cite{zuo2023missing,li2018dcrnn}. More information on datasets and experimental setups is provided in Appendix~C.

\paragraph{Zero-shot Benchmarks.}

Table~\ref{tab:horizon_results} compares GAS-PFN with PFN-based and general-purpose zero-shot forecasters across six datasets and three prediction horizons. Each baseline is evaluated under two prespecified context regimes. The recent-only regime uses the same 18 most recent observations as GAS-PFN. The history-augmented regime uses the context length specified by the baseline's default configuration, with the final 18 observations drawn from the recent window and any earlier inputs drawn from the dataset's chronological training split. This gives each baseline access to the same recent history and training-split historical data pool as GAS-PFN without imposing a common context length. Table~\ref{tab:horizon_results} reports, for each baseline and dataset, the regime with lower cumulative 18-step MAE; complete results for both regimes are provided in Appendix~D. GAS-PFN uses a fixed mixed-context configuration and the same hyperparameters across all datasets. For PEMS07, the largest sensor network, GAS-PFN uses memory-bounded channel-cohort inference; details are provided in Appendix~C.

GAS-PFN achieves the best result in 27 of 36 dataset--horizon--metric comparisons, including the lowest MSE in 14 of 18 settings. At the 90-minute horizon, it obtains the best MAE and MSE on five of six datasets. Its MSE is 34.0\% lower than TimePFN on average across the six datasets and 27.7\% lower on average across comparisons with four recent industry-developed general-purpose zero-shot forecasters. Additional experiments in Appendix~D show that GAS-PFN's advantage over TimePFN persists even when TimePFN is fine-tuned on target-domain data while GAS-PFN receives no parameter updates. GAS-PFN outperforms fine-tuned TimePFN across every dataset and horizon under a matched 32-patch historical-data budget and remains stronger in a majority of settings when TimePFN is given 64 patches for fine-tuning, twice GAS-PFN's budget. In addition, ablations in Appendix~E provide
consistent evidence that the dual-branch architecture, mixed-context support, and hybrid neighborhood routing each contribute to inference-time forecasting performance.

\begin{figure}[!t]
    \centering
    \includegraphics[
        width=\columnwidth,
    ]{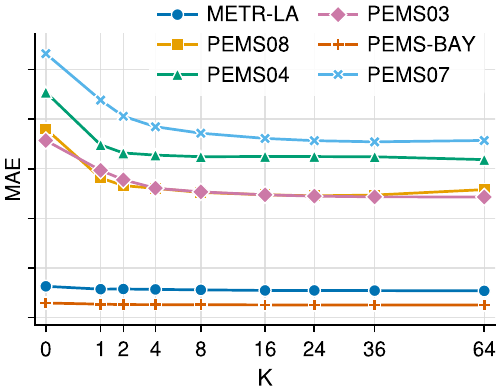}
\caption{
MAE as a function of the per-source context budget $K$ at the 90-minute horizon. Each setting uses $K$ recent and $K$ historical support patches per node. Values are computed using a single fixed GAS-PFN checkpoint.
}
    \label{fig:context_sweep}
\end{figure}

\paragraph{Context budget analysis.}
Figure~\ref{fig:context_sweep} varies the number of recent and historical support patches per node. At the 90-minute horizon, increasing the budget from no context to $K=16$ recent and $K=16$ historical patches reduces MAE by 25.4\% on average. The improvement is particularly large on PEMS03, PEMS04, PEMS07, and PEMS08, where MAE decreases by 28.5--34.9\%, but is smaller on METR-LA and PEMS-BAY, where it decreases by 12.9--13.2\%. Beyond $K=16$, gains become smaller and dataset-dependent, suggesting diminishing returns beyond 32 total support patches per node. The smaller context gains on METR-LA and PEMS-BAY coincide with less consistent advantages over the zero-shot baselines in Table~\ref{tab:horizon_results}.

\begin{figure}[!t]
    \centering
    \includegraphics[
        width=\columnwidth,
    ]{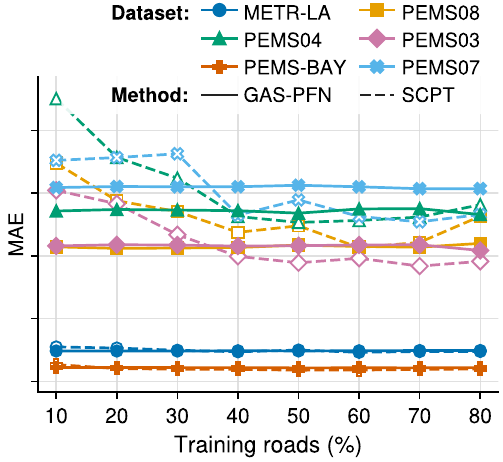}
\caption{
MAE of GAS-PFN and SCPT at the 60-minute horizon as the fraction of roads available for SCPT training varies. Values average three runs with matched road-split and checkpoint seeds.
}
    \label{fig:scpt_comparison}
\end{figure}

\paragraph{Cross-road generalization.}
Figure~\ref{fig:scpt_comparison} compares GAS-PFN with SCPT at the
12-step (60-minute) horizon across varying supervised road budgets. This comparison allows each model to operate under its own evaluation paradigm, and favors SCPT in terms of target-specific training: SCPT is trained using the full historical record from the indicated fraction of roads, while forecasts on held-out roads use its maximum lookback of 60 minutes; GAS-PFN performs zero-shot inference without dataset-specific parameter updates. Despite SCPT's access to target-specific training, GAS-PFN achieves lower MAE and MSE on all six datasets when SCPT is trained on only 10\% of the roads, reducing MSE by 34.7\% on average. GAS-PFN remains better on five of six datasets at both the 20\% and 30\% budgets, with average MSE reductions of 20.0\% and 11.1\%, respectively. On PEMS08, GAS-PFN matches or outperforms SCPT across all road-training fractions. SCPT generally becomes stronger as its supervised road coverage increases, showing that GAS-PFN is particularly effective when target-network training coverage is limited or unavailable. Complete MAE and MSE results are provided in Appendix~D.

\section{Conclusion}

We introduced GAS-PFN, a graph-aware Prior-Data Fitted Network for zero-shot traffic forecasting. GAS-PFN co-designs graph-structured synthetic pretraining tasks with mixed-context memory and dual mask-conditioned inference, enabling it to adapt to a target road network without parameter updates and with only 90 minutes of contiguous recent history. Across six traffic benchmarks, GAS-PFN achieves the strongest overall zero-shot performance. It also remains competitive with supervised cross-road forecasting
under limited target-road training. These results suggest utility in two deployment settings: forecasting on newly added roads when supervised coverage is limited, and forecasting on existing networks when recent observations are interrupted but older measurements remain available. Conceptually, these results support synthetic task priors as a practical mechanism for introducing domain-specific structure into PFNs, suggesting applications in other structured prediction domains. \noindent\textbf{Disclosure:} Generative AI tools assisted with language editing, \LaTeX{} formatting, and manuscript review.

\bibliography{ref}

%\input{ReproducibilityChecklist.tex}

\end{document}

\maketitle

% This changes numbered sections to Appendix A, Appendix B, etc.
\appendix

\section{Additional Synthetic Prior Details}
\label{app:synthetic-prior}

The synthetic prior used for pretraining was generated offline and stored
as a frozen bank. Construction proceeded in two stages. First, a bank of
univariate temporal base series was generated from compositional trend,
seasonal, and stochastic components. Each base series was then lifted to
a sampled road subgraph and modified using node-specific gains,
random-walk diffusion, and graph-correlated process noise. The resulting
bank contains one million multivariate series of length $T=12{,}000$.

\subsection{Temporal Base Generator}
\label{app:temporal-generator}

The univariate temporal component builds on the compositional synthetic
priors introduced by ForecastPFN and subsequently adapted by
LaT-PFN~\cite{dooley2023forecastpfn,verdenius2024latpfn}. GAS-PFN modifies
this temporal base and extends it with road-network sampling and
graph-structured spatial perturbations to produce multivariate traffic
processes. The precise modifications and parameter distributions are
documented below.

Let $i$ index a univariate series and
$t\in\{0,\ldots,T-1\}$ index its stored time points. Several
hyperparameters are shared by series generated together in a vectorized
generator call. We denote by $\mathcal{C}$ the index set of series
produced in one such call and state explicitly when a sampled quantity
is shared within $\mathcal{C}$. Unless noted otherwise, the underlying
random draws are independent.

\paragraph{Synthetic time coordinate.}
A temporal resolution is sampled once per generator call as
\begin{equation}
U_{\rho}
\sim
\mathcal{U}(\log_2 6,\log_2 61),
\qquad
\rho
=
\frac{2^{U_{\rho}}-1}{50}.
\label{eq:app-resolution}
\end{equation}
The corresponding coordinate span and time coordinate are
\begin{equation}
x_{\max}
=
\left\lceil\frac{T}{\rho}\right\rceil,
\qquad
x(t)
=
\frac{t}{T-1}x_{\max}.
\label{eq:app-time-coordinate}
\end{equation}
The coordinate $x(t)$ is measured in nominal days for the purpose of
constructing weekly, monthly, and annual seasonal phases. It does not
represent a real calendar timestamp.

\paragraph{Trend component.}
For each series $i\in\mathcal{C}$, the raw linear slope is sampled as
\begin{equation}
\widetilde{\lambda}^{\mathrm{lin}}_i
=
-0.015
+
0.03B^{\mathrm{lin}}_i
+
0.005\xi^{\mathrm{lin}}_i.
\label{eq:app-raw-linear-slope}
\end{equation}
Here,
$B^{\mathrm{lin}}_i\sim\operatorname{Beta}(2,2)$ and
$\xi^{\mathrm{lin}}_i\sim\mathcal{N}(0,1)$ are independent.
Slopes generated in the same call are aligned to the sign of their
median:
\begin{equation}
\lambda^{\mathrm{lin}}_i
=
\left|\widetilde{\lambda}^{\mathrm{lin}}_i\right|
\operatorname{sign}
\left(
\operatorname{median}_{i'\in\mathcal{C}}
\widetilde{\lambda}^{\mathrm{lin}}_{i'}
\right).
\label{eq:app-aligned-linear-slope}
\end{equation}
The linear offset is sampled independently for each series:
\begin{equation}
o^{\mathrm{lin}}_i
\sim
\mathcal{U}(-1,1.2).
\label{eq:app-linear-offset}
\end{equation}

For the exponential component, let
\begin{equation}
B^{\mathrm{exp}}_i
\sim
\operatorname{Beta}(2,2),
\qquad
\xi^{\mathrm{exp}}_i
\sim
\mathcal{N}(0,1).
\end{equation}
The raw exponential base is
\begin{equation}
\begin{aligned}
\widetilde{b}^{\mathrm{exp}}_i
={}&
1+300\log_2\!\Bigl[
2^{-1/60000}\\
&+
\left(
2^{1/600}-2^{-1/60000}
\right)B^{\mathrm{exp}}_i
+
0.001\xi^{\mathrm{exp}}_i
\Bigr].
\end{aligned}
\label{eq:app-raw-exp-base}
\end{equation}
As with the linear slope, deviations from one are aligned to a common
sign within the generator call:
\begin{equation}
b^{\mathrm{exp}}_i
=
1+
\left|\widetilde{b}^{\mathrm{exp}}_i-1\right|
\operatorname{sign}
\left[
\operatorname{median}_{i'\in\mathcal{C}}
\left(
\widetilde{b}^{\mathrm{exp}}_{i'}-1
\right)
\right].
\label{eq:app-aligned-exp-base}
\end{equation}
The exponential offset and lower-bounded base are
\begin{equation}
o^{\mathrm{exp}}_i
\sim
\mathcal{U}(-1.5,2),
\qquad
\overline{b}^{\mathrm{exp}}_i
=
\max\!\left(b^{\mathrm{exp}}_i,10^{-5}\right).
\label{eq:app-exp-offset}
\end{equation}

Combining the linear and exponential components gives the total trend
factor
\begin{equation}
\begin{aligned}
\mathcal{T}_i(t)
={}&
\left[
1+\lambda^{\mathrm{lin}}_i
\left(
x(t)-o^{\mathrm{lin}}_i x_{\max}
\right)
\right]\\
&\times
\left(\overline{b}^{\mathrm{exp}}_i\right)^{
x(t)/x_{\max}-o^{\mathrm{exp}}_i}.
\end{aligned}
\label{eq:app-total-trend}
\end{equation}
The lower bound in Eq.~\eqref{eq:app-exp-offset} applies only to the base
of the exponential component; the resulting trend and final temporal
series are not clipped.

\paragraph{Seasonal component.}
Let
\begin{equation}
\mathcal{J}
=
\{\mathrm{yr},\mathrm{mo},\mathrm{wk}\}
\end{equation}
index the annual, monthly, and weekly seasonal families. For a cycle
length $q>0$ and bin width $p>0$, define
\begin{equation}
\iota_{q,p}(x)
=
\left\lfloor\frac{x}{p}\right\rfloor+1
\end{equation}
and
\begin{equation}
\operatorname{bin}_{q,p}(x)
=
\begin{cases}
\iota_{q,p}(x),
& \iota_{q,p}(x)\le q,\\[1mm]
\bigl(\iota_{q,p}(x)\bmod q\bigr)+1,
& \iota_{q,p}(x)>q.
\end{cases}
\label{eq:app-binning}
\end{equation}
The seasonal phases are
\begin{equation}
\begin{aligned}
\phi_{\mathrm{yr}}(t)
&=
\frac{\operatorname{bin}_{12,30.417}(x(t))}{12},\\
\phi_{\mathrm{mo}}(t)
&=
\frac{\operatorname{bin}_{30.417,1}(x(t))}{30.417},\\
\phi_{\mathrm{wk}}(t)
&=
\frac{\operatorname{bin}_{7,1}(x(t))}{7}.
\end{aligned}
\label{eq:app-seasonal-phases}
\end{equation}

For each series, the corresponding seasonal amplitudes are
\begin{equation}
\begin{aligned}
\omega_{i,\mathrm{yr}}
&=
-12+24B_{i,\mathrm{yr}}
+0.15\xi^{\mathrm{sea}}_{i,\mathrm{yr}},\\
\omega_{i,\mathrm{mo}}
&=
-6+12B_{i,\mathrm{mo}}
+0.15\xi^{\mathrm{sea}}_{i,\mathrm{mo}},\\
\omega_{i,\mathrm{wk}}
&=
-3+6B_{i,\mathrm{wk}}
+0.15\xi^{\mathrm{sea}}_{i,\mathrm{wk}},
\end{aligned}
\label{eq:app-seasonal-amplitudes}
\end{equation}
where
\begin{equation}
B_{i,j}\sim\operatorname{Beta}(2,2),
\qquad
\xi^{\mathrm{sea}}_{i,j}\sim\mathcal{N}(0,1),
\qquad
j\in\mathcal{J},
\end{equation}
with independent draws across series and seasonal families. Because
Gaussian perturbations are added after the Beta-scaled draws, the
displayed endpoint values are not strict support bounds.

The number of harmonics for each seasonal family is sampled
independently and uniformly:
\begin{equation}
H_j
\overset{\mathrm{iid}}{\sim}
\operatorname{DiscreteUniform}
\{2,3,4,5,6,7\},
\qquad
j\in\mathcal{J}.
\label{eq:app-harmonic-count}
\end{equation}
Each $H_j$ is shared by the series in the same generator call. For
series $i$, family $j$, and harmonic $h\in\{1,\ldots,H_j\}$, raw sine
and cosine coefficients are sampled independently as
\begin{equation}
\widetilde{a}_{ijh}
\sim
\mathcal{N}(0,h^{-2}),
\qquad
\widetilde{c}_{ijh}
\sim
\mathcal{N}(0,h^{-2}).
\label{eq:app-raw-fourier-coefficients}
\end{equation}
The coefficients for each seasonal family are normalized jointly over
the series and harmonic dimensions of the generator call. Define
\begin{equation}
\kappa_j
=
\left[
\sum_{i\in\mathcal{C}}
\sum_{h=1}^{H_j}
\left(
\widetilde{a}_{ijh}^{\,2}
+
\widetilde{c}_{ijh}^{\,2}
\right)
\right]^{1/2}.
\label{eq:app-fourier-normalizer}
\end{equation}
The normalized coefficients are
\begin{equation}
a_{ijh}
=
\frac{\widetilde{a}_{ijh}}{\kappa_j},
\qquad
c_{ijh}
=
\frac{\widetilde{c}_{ijh}}{\kappa_j}.
\label{eq:app-normalized-fourier-coefficients}
\end{equation}

The Fourier component for seasonal family $j$ is
\begin{equation}
F_{ij}(t)
=
\sum_{h=1}^{H_j}
\left[
a_{ijh}\sin\!\left(2\pi h\phi_j(t)\right)
+
c_{ijh}\cos\!\left(2\pi h\phi_j(t)\right)
\right].
\label{eq:app-fourier-component}
\end{equation}
The complete seasonal factor is
\begin{equation}
s_i(t)
=
\prod_{j\in\mathcal{J}}
\left[
1+\omega_{i,j}F_{ij}(t)
\right].
\label{eq:app-total-seasonality}
\end{equation}
No clipping or positivity constraint is imposed on the individual
seasonal factors or their product.

\paragraph{Stochastic component and composition.}
The Weibull shape parameter is sampled independently for each series:
\begin{equation}
k_i
=
0.5+3B^{\mathrm{wei}}_i,
\qquad
B^{\mathrm{wei}}_i
\sim
\operatorname{Beta}(2,2).
\label{eq:app-weibull-shape}
\end{equation}
Once per generator call, a noise-amplitude interval is selected as
\begin{equation}
(L_{\nu},U_{\nu})
=
\begin{cases}
(0,0.2),
& \text{with probability }0.4,\\
(0.3,0.7),
& \text{with probability }0.4,\\
(0.8,1.2),
& \text{with probability }0.2.
\end{cases}
\label{eq:app-noise-band}
\end{equation}
Conditional on this shared interval, the series-level noise amplitude is
sampled independently as
\begin{equation}
\nu_i
\sim
\mathcal{U}(L_{\nu},U_{\nu}).
\label{eq:app-noise-amplitude}
\end{equation}

Let
\begin{equation}
R_i(t)
\sim
\operatorname{Weibull}(k_i,1)
\label{eq:app-weibull-draw}
\end{equation}
denote a unit-scale Weibull realization sampled independently over
series and time. Its median-rescaled form is
\begin{equation}
W_i(t)
=
(\log 2)^{-1/k_i}R_i(t),
\label{eq:app-median-scaled-weibull}
\end{equation}
for which $\operatorname{median}[W_i(t)]=1$. The resulting
multiplicative stochastic factor is
\begin{equation}
m_i(t)
=
1+\nu_i\left[W_i(t)-1\right].
\label{eq:app-stochastic-factor}
\end{equation}

An overall amplitude is sampled independently for each series:
\begin{equation}
\zeta_i
\sim
\mathcal{U}(0.5,4).
\label{eq:app-overall-amplitude}
\end{equation}
The composition branch and additive-branch trend weight are sampled once
per generator call:
\begin{equation}
\chi_{\mathrm{add}}
\sim
\operatorname{Bernoulli}(0.5),
\qquad
w_{\mathrm{add}}
\sim
\mathcal{U}(0.2,0.3),
\label{eq:app-composition-parameters}
\end{equation}
where $\chi_{\mathrm{add}}=1$ selects additive composition and
$\chi_{\mathrm{add}}=0$ selects multiplicative composition.

The stored univariate temporal base is
\begin{equation}
y_{0,i}(t)
=
\begin{cases}
\zeta_i
\left[
w_{\mathrm{add}}\mathcal{T}_i(t)+s_i(t)
\right]
\mathcal{T}_i(t)m_i(t),
& \chi_{\mathrm{add}}=1,\\[2mm]
\zeta_i
\mathcal{T}_i(t)^2s_i(t)m_i(t),
& \chi_{\mathrm{add}}=0.
\end{cases}
\label{eq:app-temporal-base}
\end{equation}
Equation~\eqref{eq:app-temporal-base} makes the final trend
multiplication explicit: after the additive or multiplicative
composition is formed, it is multiplied by $\mathcal{T}_i(t)$ and
$m_i(t)$. The multiplicative branch therefore contains
$\mathcal{T}_i(t)^2$. No normalization,
standardization, clipping, discretization, zero inflation,
non-negativity transformation, or acceptance--rejection step is applied
to the resulting temporal base series.

\subsection{Road-Graph Bank and Subgraph Sampling}
\label{app:road-graph-bank}

\paragraph{Graph-bank construction procedure.}
The road-graph bank was constructed from OpenStreetMap drive networks
obtained through OSMnx. It contains networks for 20 U.S. metropolitan
areas: Atlanta, Austin, Boston, Chicago, Dallas, Denver, Houston, Los
Angeles, Miami, Minneapolis, New York, Philadelphia, Phoenix, Portland,
Sacramento, San Diego, San Francisco, San Jose, Seattle, and
Washington, DC.

For each metropolitan area, the road topology was simplified and its
largest weakly connected component was retained. Networks containing
more than 512 nodes were reduced to 512 nodes by breadth-first expansion
from a maximum-degree node in an undirected view of the graph.

The OpenStreetMap data accessed through OSMnx are used only to define
road connectivity and direction; no traffic measurements or
spatiotemporal observations from the benchmark datasets enter
synthetic-prior generation. For a sampled graph
with $N$ nodes, let
$\mathbf{A}_d\in\{0,1\}^{N\times N}$ denote its directed adjacency
matrix. For nodes $v$ and $v'$,
\begin{equation}
(\mathbf{A}_d)_{vv'}
=
\mathbb{I}\!\left[v\rightarrow v'\right],
\label{eq:app-directed-adjacency}
\end{equation}
where $\mathbb{I}[\cdot]$ is the indicator function and
$v\rightarrow v'$ denotes a directed road edge. Parallel edges between
the same ordered pair are collapsed to one binary entry. The
corresponding undirected adjacency is
\begin{equation}
\mathbf{A}_u
=
\mathbb{I}\!\left[
\mathbf{A}_d+\mathbf{A}_d^{\top}>0
\right],
\label{eq:app-undirected-adjacency}
\end{equation}
where the indicator is applied elementwise. Road length, geographic
distance, speed, road class, geometry, coordinates, and other edge
attributes are not used as adjacency or diffusion weights.

\paragraph{Subgraph-sampling procedure.}
The multivariate bank is divided into 50 shards. Target subgraph sizes
cycle through 24, 36, 48, and 60 nodes across successive shards. For
each shard, one eligible metropolitan network is selected uniformly. A
connected component large enough to contain the target number of nodes
is selected, a root node is sampled within that component, and a
connected subgraph of the target size is extracted by breadth-first
expansion with randomized neighbor order. The 50 shards use seeds
42--91 for network and subgraph sampling. Metropolitan networks are
sampled independently across shards, so repeated selections are
allowed.

\paragraph{Realized graph bank.}
One sampled subgraph is fixed for each shard and shared by all 20,000
multivariate series in that shard. The frozen bank used in the reported
experiments therefore contains 50 fixed graph configurations: 13 with
24 nodes, 13 with 36 nodes, 12 with 48 nodes, and 12 with 60 nodes.

\subsection{Graph-Structured Multivariate Process}
\label{app:graph-process}

Fix a temporal base series $y_0(t)$ generated by
Eq.~\eqref{eq:app-temporal-base} and a sampled graph $G=(V,E)$, where
$V$ and $E$ are its node and edge sets and $N=|V|$. We omit the
temporal-series index $i$ in the remainder of this section.

For $r\in\{u,d\}$, where $u$ and $d$ denote the undirected and directed
graphs, respectively, define
\begin{equation}
\begin{aligned}
\widetilde{\mathbf{A}}_r
&=
\mathbf{A}_r+\mathbf{I}_N,\\
\mathbf{D}_r
&=
\operatorname{diag}
\left(
\widetilde{\mathbf{A}}_r\mathbf{1}_N
\right),\\
\mathbf{P}_r
&=
\mathbf{D}_r^{-1}\widetilde{\mathbf{A}}_r.
\end{aligned}
\label{eq:app-random-walk-matrices}
\end{equation}
Here, $\mathbf{I}_N$ is the $N\times N$ identity matrix,
$\mathbf{1}_N$ is the $N$-dimensional all-ones vector, and
$\mathbf{0}_N$ denotes the $N$-dimensional zero vector. The sampled
subgraphs used in the reported frozen bank contained no diagonal
adjacency entries before $\mathbf{I}_N$ was added, so each node receives
one explicit self-loop.

Both $\mathbf{P}_u$ and $\mathbf{P}_d$ are row-stochastic random-walk
matrices. Under the directed convention in
Eq.~\eqref{eq:app-directed-adjacency}, row $v$ of $\mathbf{P}_d$
averages node $v$ with its outgoing road neighbors. For a graph signal
$\mathbf{x}\in\mathbb{R}^{N}$, define the one-step lazy random-walk
operator
\begin{equation}
\mathcal{P}_r(\mathbf{x})
=
\frac{1}{2}
\left(
\mathbf{x}+\mathbf{P}_r\mathbf{x}
\right),
\qquad
r\in\{u,d\}.
\label{eq:app-lazy-random-walk}
\end{equation}
The graph-diffusion operator $\mathcal{P}$ used in the main-paper overview is the undirected operator $\mathcal{P}_u$ used here.

The temporal base is replicated across graph nodes as
\begin{equation}
\mathbf{y}_0(t)
=
y_0(t)\mathbf{1}_N.
\label{eq:app-replicated-temporal-base}
\end{equation}
Node-specific gains are sampled once per multivariate series and remain
fixed over time:
\begin{equation}
\begin{aligned}
\boldsymbol{\xi}_g
&\sim
\mathcal{N}(\mathbf{0}_N,\mathbf{I}_N),\\
\mathbf{g}
&=
\mathbf{1}_N+\sigma_g\boldsymbol{\xi}_g,\\
\mathbf{z}(t)
&=
\mathbf{g}\odot\mathbf{y}_0(t),
\end{aligned}
\label{eq:app-node-gains}
\end{equation}
where $\odot$ denotes elementwise multiplication and
$\sigma_g=0.05$.

The deterministic graph-propagated signal is
\begin{equation}
\mathbf{z}_{\mathrm{prop}}(t)
=
(1-\beta)\mathbf{z}(t)
+
\beta\mathcal{P}_u\!\left(\mathbf{z}(t)\right),
\qquad
\beta=0.7.
\label{eq:app-deterministic-propagation}
\end{equation}
Equivalently,
\begin{equation}
\mathbf{z}_{\mathrm{prop}}(t)
=
0.65\mathbf{z}(t)
+
0.35\mathbf{P}_u\mathbf{z}(t).
\label{eq:app-deterministic-propagation-expanded}
\end{equation}
Thus, deterministic propagation of the temporal signal uses only the
undirected road topology. The remaining weight $1-\beta=0.3$ is an
identity contribution, not a directed-transport term.

For the process-noise component, a Gaussian vector is sampled
independently at each time and independently of the node gains:
\begin{equation}
\mathbf{n}(t)
\sim
\mathcal{N}(\mathbf{0}_N,\mathbf{I}_N).
\label{eq:app-process-noise}
\end{equation}
The same realization is filtered through both graph operators.
Following the notation in the main paper, we absorb the
process-noise scale into $\boldsymbol{\epsilon}_G(t)$:
\begin{equation}
\begin{aligned}
\boldsymbol{\epsilon}_G(t)
&=
\sigma_{\epsilon}
\Big[
\beta\mathcal{P}_u\!\left(\mathbf{n}(t)\right)
+
(1-\beta)\mathcal{P}_d\!\left(\mathbf{n}(t)\right)
\Big], \\
\sigma_{\epsilon} &= 0.02 .
\end{aligned}
\label{eq:app-graph-correlated-noise}
\end{equation}
The final multivariate signal is
\begin{equation}
\begin{aligned}
\mathbf{y}(t)
&=
\mathbf{y}_0(t)
+
\alpha
\Big[
\mathbf{z}_{\mathrm{prop}}(t)
-
\mathbf{y}_0(t)
+
\boldsymbol{\epsilon}_G(t)
\Big], \\
\alpha &= 0.05 .
\end{aligned}
\label{eq:app-final-multivariate-signal}
\end{equation}
For reference, the complete process-noise contribution after applying
the outer scale $\alpha$ is
\begin{equation}
\begin{aligned}
\alpha\boldsymbol{\epsilon}_G(t)
={}&
0.0005\mathbf{n}(t)
+
0.00035\mathbf{P}_u\mathbf{n}(t)\\
&+
0.00015\mathbf{P}_d\mathbf{n}(t).
\end{aligned}
\label{eq:app-expanded-process-noise}
\end{equation}
The directed adjacency enters through the graph-correlated process-noise
component, whereas deterministic propagation uses the undirected road
topology.

\section{Additional Method Details}
\label{app:method-details}

This section provides additional details on pretraining-task sampling,
context construction, tokenization, neighborhood routing, attention
masking, and the training objective. The synthetic spatiotemporal prior
is described separately in Appendix~\ref{app:synthetic-prior}.

\subsection{Pretraining Task Construction}
\label{app:pretraining-tasks}

Each pretraining task begins with one multivariate realization
$\{\mathbf{y}(t)\}_{t=0}^{T-1}$ and its associated graph
$G=(V,E)$, where $\mathbf{y}(t)\in\mathbb{R}^{|V|}$ and
$T=12{,}000$. The lookback and forecast lengths are fixed to
\begin{equation}
L=H=18.
\end{equation}

\paragraph{Temporal partition.}
Let
\begin{equation}
T_{\mathrm{split}}
=
\left\lfloor 0.8T\right\rfloor.
\end{equation}
Eligible context and query endpoints are
\begin{equation}
\begin{aligned}
\mathcal{E}_{\mathrm{ctx}}
=
\{\,\tau:
&\; L-1\leq\tau\leq T-H-1, \\
&\; \tau+H\leq T_{\mathrm{split}}
\,\},
\end{aligned}
\label{eq:app-context-endpoints}
\end{equation}
and
\begin{equation}
\begin{aligned}
\mathcal{E}_{\mathrm{qry}}
=
\{\,\tau:
&\; L-1\leq\tau\leq T-H-1, \\
&\; \tau-L+1\geq T_{\mathrm{split}}
\,\}.
\end{aligned}
\label{eq:app-query-endpoints}
\end{equation}
Every context target therefore ends at or before the temporal split,
while every query lookback begins at or after it.

\paragraph{Context and query sampling.}
The number of raw context patches is sampled as
\begin{equation}
C
=
\left\lfloor
\exp(U_C)
\right\rfloor,
\qquad
U_C
\sim
\mathcal{U}
\left(
\log 64,\log 768
\right),
\label{eq:app-context-count}
\end{equation}
and the number of query patches is
\begin{equation}
Q
\sim
\operatorname{DiscreteUniform}
\{1,\ldots,32\}.
\label{eq:app-query-count}
\end{equation}

For each context occurrence $r\in\{1,\ldots,C\}$, an endpoint
$\tau_r^{\mathrm{ctx}}$ is sampled uniformly from
$\mathcal{E}_{\mathrm{ctx}}$ and a node
$v_r^{\mathrm{ctx}}$ is sampled uniformly from $V$. Query occurrences
are sampled analogously from $\mathcal{E}_{\mathrm{qry}}$ and $V$.
Sampling is performed with replacement.

For each node $v$, let
\begin{equation}
\mathcal{C}^{\mathrm{raw}}_v
=
\left\{
\tau_r^{\mathrm{ctx}}:
v_r^{\mathrm{ctx}}=v
\right\}
\end{equation}
denote its raw context-endpoint collection, and let
\begin{equation}
\mathcal{Q}_v
=
\left\{
\tau_r^{\mathrm{qry}}:
v_r^{\mathrm{qry}}=v
\right\}
\end{equation}
denote its query-endpoint collection. These collections are treated as
multisets during pretraining because duplicate node--endpoint draws are
possible. A task need not contain every node in the stored graph. The
task-local graph is restricted to the distinct nodes represented by at
least one context or query occurrence.

For a node $v$ and endpoint $\tau$, the node-local lookback and target
patches are
\begin{equation}
\mathbf{x}_{v,\tau}
=
\left[
y_v(\tau-L+1),\ldots,y_v(\tau)
\right]
\in\mathbb{R}^{L},
\label{eq:app-lookback-patch}
\end{equation}
and
\begin{equation}
\mathbf{y}_{v,\tau}
=
\left[
y_v(\tau+1),\ldots,y_v(\tau+H)
\right]
\in\mathbb{R}^{H}.
\label{eq:app-target-patch}
\end{equation}
Overlapping windows and exact duplicate patches are permitted.

\paragraph{Task-level scaling.}
A shared normalizer is computed from all scalar values in the sampled
context lookbacks. Let $\mu_{\mathrm{ctx}}$ and
$m_{\mathrm{ctx}}$ denote their mean and median, respectively, and define
\begin{equation}
s_{\mathrm{ctx}}
=
\max\left\{
1.4826\,
\operatorname{median}_{r,\ell}
\left|x_{r\ell}-m_{\mathrm{ctx}}\right|,
0.10
\right\}.
\label{eq:app-task-scale}
\end{equation}
Every context and query lookback and target is transformed elementwise as
\begin{equation}
\mathcal{N}_{\mathrm{task}}(\mathbf{a})
=
\operatorname{clip}
\left(
\frac{\mathbf{a}-\mu_{\mathrm{ctx}}}{s_{\mathrm{ctx}}},
-10,10
\right).
\label{eq:app-task-normalization}
\end{equation}
No missing-value or zero-target mask is synthesized during pretraining.
\subsection{Tokenization and Sequence Assembly}
\label{app:tokenization}

Let $\overline{\mathbf{x}}_{v,\tau}$ and
$\overline{\mathbf{y}}_{v,\tau}$ denote the normalized lookback and
target patches under the active preprocessing pipeline. During
pretraining, these are obtained using
Eq.~\eqref{eq:app-task-normalization}; benchmark preprocessing is
described in Appendix~\ref{app:dataset-details}.

Context and query patches are embedded by separate affine projections:
\begin{equation}
\mathbf{e}^{\mathrm{ctx}}_{v,\tau}
=
\mathbf{W}_{\mathrm{ctx}}
\left[
\overline{\mathbf{x}}_{v,\tau};
\overline{\mathbf{y}}_{v,\tau}
\right]
+
\mathbf{b}_{\mathrm{ctx}},
\label{eq:app-context-embedding}
\end{equation}
and
\begin{equation}
\mathbf{e}^{\mathrm{qry}}_{v,\tau}
=
\mathbf{W}_{\mathrm{qry}}
\left[
\overline{\mathbf{x}}_{v,\tau};
\mathbf{0}_{H}
\right]
+
\mathbf{b}_{\mathrm{qry}},
\label{eq:app-query-embedding}
\end{equation}
where
\begin{equation}
\mathbf{W}_{\mathrm{ctx}},
\mathbf{W}_{\mathrm{qry}}
\in
\mathbb{R}^{d\times(L+H)},
\qquad
\mathbf{b}_{\mathrm{ctx}},
\mathbf{b}_{\mathrm{qry}}
\in
\mathbb{R}^{d},
\end{equation}
$d=512$, and $\mathbf{0}_{H}$ is the $H$-dimensional zero vector. The
two projections have independent parameters and contain no additional
normalization, activation, or hidden layer.

Each token retains its node $v$, endpoint $\tau$, and context/query
role as separate metadata. These metadata are used to construct
attention masks, temporal attention terms, and routed neighborhoods;
they are not added to the token value. The architecture contains no
node embedding, timestamp embedding, context/query type embedding, or
absolute sequence-position embedding.

After context construction, let
$\widetilde{\mathcal{C}}_v$ denote the retained context endpoints for
node $v$. The Transformer input is assembled as
\begin{equation}
\mathbf{Z}^{(0)}
=
\left[
\left(
\mathbf{e}^{\mathrm{ctx}}_{v,\tau}
\right)_{
v\in V_{\mathrm{tok}},
\,\tau\in\widetilde{\mathcal{C}}_v
};
\left(
\mathbf{e}^{\mathrm{qry}}_{v,\tau}
\right)_{
v\in V_{\mathrm{tok}},
\,\tau\in\mathcal{Q}_v
}
\right],
\label{eq:app-transformer-input}
\end{equation}
where
\begin{equation}
V_{\mathrm{tok}}
=
\left\{
v\in V:
\widetilde{\mathcal{C}}_v\neq\varnothing
\ \text{or}\
\mathcal{Q}_v\neq\varnothing
\right\}.
\end{equation}
All retained context occurrences precede all query occurrences. Context
tokens are globally ordered by endpoint before the query block is
appended. Chronology enters the model through endpoint metadata and
attention constraints rather than absolute sequence position.

\subsection{Context Construction}
\label{app:context-construction}

The reported checkpoints and target-dataset evaluations use different
context-assembly procedures. During pretraining, a larger randomly
sampled context is reduced using recent raw tokens, historical raw
tokens, and attention-pooled memory tokens. During reported zero-shot
evaluation, the pooling module is bypassed and each node receives
$K=16$ recent and $K=16$ historical raw support patches.

\subsubsection{Pretraining Context Construction}
\label{app:pretraining-context}

Each pretraining task initially samples between 64 and 768 context
patches. The context is capped at 16 tokens per represented sensor,
comprising up to eight recent raw patches, four older raw patches, and
four pooled summaries of the remaining older context. Sensors with fewer
sampled patches contribute fewer tokens. The resulting retained context
endpoints are denoted by $\widetilde{\mathcal{C}}_v$.

At target-dataset inference, GAS-PFN instead uses the fixed raw-support
context described below.

\subsubsection{Target-Dataset Mixed Context}
\label{app:target-context}

Let
\begin{equation}
\left\{
\tau_u
\right\}_{u=0}^{W-1}
\end{equation}
denote the endpoints of all stride-one windows from the target
dataset's chronological training split, ordered by time. The
recent endpoint set is
\begin{equation}
\mathcal{C}^{\mathrm{recent}}
=
\left\{
\tau_{W-K},\ldots,\tau_{W-1}
\right\},
\qquad
K=16.
\label{eq:app-recent-support}
\end{equation}
The same recent endpoints are used for every node, preserving temporal
alignment across sensors.

Historical anchors are drawn from
\begin{equation}
\mathcal{F}
=
\left\{
0,\ldots,W-K-1
\right\}.
\end{equation}
To prefer windows with informative targets, define
\begin{equation}
r_u
=
\frac{1}{|V|H}
\sum_{v\in V}
\sum_{h=1}^{H}
\mathbb{I}
\left[
\left(
\mathbf{y}^{\mathrm{raw}}_{v,u}
\right)_h
\neq 0
\right],
\label{eq:app-window-validity}
\end{equation}
where $\mathbf{y}^{\mathrm{raw}}_{v,u}$ is the original-scale target
associated with training window $u$, $(\cdot)_h$ selects forecast step
$h$, and $\mathbb{I}[\cdot]$ denotes the indicator function. The
preferred historical indices are
\begin{equation}
\mathcal{F}^{+}
=
\left\{
u\in\mathcal{F}:r_u\geq 0.5
\right\}.
\end{equation}

For each node $v$, eight anchors are first drawn across equal-count
chronological strata of $\mathcal{F}^{+}$. If fewer than eight preferred
indices are available, all available preferred indices are retained and
the remaining positions are filled by stratified sampling from
$\mathcal{F}$. After removing the first group, eight additional anchors
are drawn across equal-count chronological strata of the remaining
historical indices.

Let
\begin{equation}
\mathcal{C}^{\mathrm{anchor}}_v
=
\left\{
\tau_u:
u\text{ is selected as an anchor for }v
\right\}.
\end{equation}
Anchor selection is performed independently for each node, so historical
anchors need not align across sensors. Selected anchor indices are
unique within a node, although their lookback or target windows may
overlap.

The retained target-domain context endpoints are
\begin{equation}
\widetilde{\mathcal{C}}_v
=
\mathcal{C}^{\mathrm{recent}}
\cup
\mathcal{C}^{\mathrm{anchor}}_v,
\qquad
\left|
\widetilde{\mathcal{C}}_v
\right|
=
32.
\label{eq:app-target-context-set}
\end{equation}
Historical anchors retain their original training-window endpoints. The
context is selected once and reused for all test queries and all three
reported checkpoints. No attention-pooled memory tokens are supplied
during target-dataset evaluation.

\subsection{Graph-Representation Neighborhood Routing}
\label{app:neighborhood-routing}

The multi-channel branch uses a hard router that combines target-graph
connectivity with content-derived representation similarity. Let
$\mathbf{A}\in\mathbb{R}^{|V_{\mathrm{tok}}|\times
|V_{\mathrm{tok}}|}$ denote the graph-score matrix restricted to
$V_{\mathrm{tok}}$. Before routing, its diagonal is set to one. The
matrix is not otherwise symmetrized, binarized, or degree-normalized;
its stored directionality and weights are retained.

Each graph-score row is independently min--max scaled:
\begin{equation}
\overline{A}_{vv'}
=
\frac{
A_{vv'}-\displaystyle\min_{w\in V_{\mathrm{tok}}}A_{vw}
}{
\displaystyle\max_{w\in V_{\mathrm{tok}}}A_{vw}
-
\displaystyle\min_{w\in V_{\mathrm{tok}}}A_{vw}
+
10^{-8}
}.
\label{eq:app-row-scaled-graph}
\end{equation}

\paragraph{Content-derived representations.}
A fixed, seed-initialized MLP
$\psi_{\omega_0}:\mathbb{R}^{d}\rightarrow\mathbb{R}^{d}$ is applied
to the learned patch-token embeddings:
\begin{equation}
\begin{aligned}
\psi_{\omega_0}(\mathbf{z})
&=
\mathbf{W}_2
\operatorname{Drop}_{0.1}
\!\left[
\operatorname{GELU}
\!\left(
\mathbf{W}_1
\operatorname{LN}(\mathbf{z})
+
\mathbf{b}_1
\right)
\right]
+
\mathbf{b}_2 .
\end{aligned}
\label{eq:app-routing-encoder}
\end{equation}
where $\operatorname{LN}$ denotes layer normalization,
$\mathbf{W}_1,\mathbf{W}_2\in\mathbb{R}^{d\times d}$, and
$\mathbf{b}_1,\mathbf{b}_2\in\mathbb{R}^{d}$. Dropout is disabled
during evaluation.

For each represented node, define the multiset of its context and query
endpoints as
\begin{equation}
\mathcal{T}_v
=
\widetilde{\mathcal{C}}_v
\uplus
\mathcal{Q}_v,
\end{equation}
where $\uplus$ denotes multiset union. Its encoded summary is
\begin{equation}
\mathbf{h}_v
=
\frac{1}{|\mathcal{T}_v|}
\sum_{\tau\in\mathcal{T}_v}
\psi_{\omega_0}
\left(
\mathbf{Z}^{(0)}_{v,\tau}
\right).
\label{eq:app-channel-summary}
\end{equation}
When multiple token occurrences share the same $(v,\tau)$ metadata,
each occurrence contributes separately to the sum.

The summary is normalized as
\begin{equation}
\widehat{\mathbf{h}}_v
=
\frac{
\mathbf{h}_v
}{
\max
\left(
\|\mathbf{h}_v\|_2,
10^{-12}
\right)
},
\label{eq:app-normalized-channel-summary}
\end{equation}
and semantic similarity is
\begin{equation}
S^{\mathrm{sem}}_{vv'}
=
\widehat{\mathbf{h}}_v^\top
\widehat{\mathbf{h}}_{v'}.
\label{eq:app-representation-similarity}
\end{equation}

The hybrid routing score is
\begin{equation}
S^{\mathrm{hyb}}_{vv'}
=
(1-\eta)\overline{A}_{vv'}
+
\eta S^{\mathrm{sem}}_{vv'},
\qquad
\eta=0.4.
\label{eq:app-hybrid-routing-score}
\end{equation}
For each node $v$, the routed neighborhood is
\begin{equation}
\mathcal{N}_v
=
\operatorname{TopK}_{v'}
\left(
S^{\mathrm{hyb}}_{vv'},
k_v
\right),
\qquad
k_v
=
\min
\left(
7,
|V_{\mathrm{tok}}|
\right).
\label{eq:app-routing-neighborhood}
\end{equation}
Selection is performed independently by row and is not symmetrized.
Self-membership is determined by the score rather than imposed
separately.

The hard TopK indices are converted directly into a Boolean attention
mask; the continuous routing scores are not passed to the Transformer.
The routing path therefore provides no forecasting-loss gradient to
$\psi_{\omega_0}$ or to its inputs, and $\omega_0$ remains at its
seed-specific initialization. The patch-token embeddings supplied to
the router are nevertheless learned through the ordinary forecasting
path, so the resulting neighborhoods remain content-dependent and may
change during pretraining. This construction avoids requiring a
continuous relaxation or surrogate-gradient estimator for discrete
neighborhood selection.

Neighborhoods are recomputed once for each branch forward from the
current token sequence. The resulting Boolean neighborhood mask is
shared across all Transformer layers.

\subsection{Exact Attention Constraints}
\label{app:attention-constraints}

A token is identified by its node and endpoint pair $(v,\tau)$ together
with its position in the token multiset. Duplicate occurrences sharing
the same $(v,\tau)$ metadata obey the same attention conditions but
remain separate sequence elements. Context and query endpoint
collections are disjoint in the executed pretraining and evaluation
protocols.

Let
\begin{equation}
\mathcal{A}^{(b)}_{(v,\tau),(v',\tau')}
\in
\{0,1\}
\end{equation}
indicate whether token $(v,\tau)$ may attend to key/value token
$(v',\tau')$ in branch
$b\in\{\mathrm{self},\mathrm{multi}\}$.

\paragraph{Self branch.}
The self-branch mask is
\begin{equation}
\mathcal{A}^{(\mathrm{self})}_{(v,\tau),(v',\tau')}
=
\mathbb{I}[v=v']\,
\mathbb{I}
\left[
\tau'\in\widetilde{\mathcal{C}}_{v'}
\right]\,
\mathbb{I}[\tau'\leq\tau].
\label{eq:app-self-mask}
\end{equation}
Both context and query tokens therefore attend only to causal context
from the same node. Context-to-query and query-to-query attention are
disabled, including query-token self-attention.

\paragraph{Multi-channel branch.}
For the reported target-dataset evaluation,
\begin{equation}
\begin{aligned}
&\mathcal{A}^{(\mathrm{multi})}_{(v,\tau),(v',\tau')}
=
\mathbb{I}
\left[
v'\in\mathcal{N}_v
\right]\\
&\quad\times
\begin{cases}
\mathbb{I}[\tau'\leq\tau],
&
\tau'\in\widetilde{\mathcal{C}}_{v'},\\[1mm]
\mathbb{I}
\left[
\tau\in\mathcal{Q}_v
\right]
\mathbb{I}[\tau'=\tau],
&
\tau'\in\mathcal{Q}_{v'}.
\end{cases}
\end{aligned}
\label{eq:app-multi-mask}
\end{equation}
Context-to-query attention remains blocked. Query-to-query attention is
permitted only among selected nodes at the same forecast origin. Here,
``same forecast time'' means equality of query-window endpoints; one
query token predicts all $H$ future steps.

During checkpoint pretraining, the query-key case instead uses
\begin{equation}
\mathbb{I}
\left[
\tau\in\mathcal{Q}_v
\right]
\mathbb{I}[\tau'\leq\tau],
\qquad
\tau'\in\mathcal{Q}_{v'},
\end{equation}
allowing a query token to attend to same-time and earlier query tokens
inside its selected neighborhood. The self branch uses
Eq.~\eqref{eq:app-self-mask} during both pretraining and evaluation.

\paragraph{Attention logits.}
At a given Transformer layer and attention head $h$, let
$\mathbf{q}^{(b,h)}_{v,\tau}$ and
$\mathbf{k}^{(b,h)}_{v',\tau'}$ denote the query and key vectors for
tokens $(v,\tau)$ and $(v',\tau')$. The layer index is omitted for
clarity. With hidden dimension $d=512$ and eight heads, the head
dimension is
\begin{equation}
d_h=64.
\end{equation}
The unmasked attention score is
\begin{equation}
\begin{aligned}
\widetilde{\ell}^{(b,h)}_{(v,\tau),(v',\tau')}
&=
\frac{
\left(\mathbf{q}^{(b,h)}_{v,\tau}\right)^\top
\mathbf{k}^{(b,h)}_{v',\tau'}
}{
\sqrt{d_h}
}\\
&\quad
-\gamma_h\max(\tau-\tau',0)
\\
&\quad
+b_{\mathrm{same},h}^{(b)}
\mathbb{I}[v=v'].
\end{aligned}
\label{eq:app-unmasked-attention-logit}
\end{equation}
The attention mask is then applied as
\begin{equation}
\ell^{(b,h)}_{(v,\tau),(v',\tau')}
=
\begin{cases}
\widetilde{\ell}^{(b,h)}_{(v,\tau),(v',\tau')},
&
\mathcal{A}^{(b)}_{(v,\tau),(v',\tau')}=1,
\\
-\infty,
&
\text{otherwise}.
\end{cases}
\label{eq:app-attention-logit}
\end{equation}
Each head learns its own temporal-distance coefficient $\gamma_h$,
initialized to
\begin{equation}
\gamma_h=0.01.
\end{equation}
The coefficients are shared across the two branches and all Transformer
layers. The self-channel coefficient
$b_{\mathrm{same},h}^{(b)}$ is learned separately for each head and
branch and is shared across layers. In the self branch, the
self-channel term is constant over every visible key and therefore
cancels under the softmax. In the multi-channel branch, it distinguishes
same-node from neighboring-node interactions.

The two branches otherwise use the same patch projections, Transformer
backbone, query/key/value projections, layer normalizations, and
feed-forward parameters.

\subsection{Fusion and Pretraining Objective}
\label{app:fusion-objective}

For a query token $(v,\tau)$ with $\tau\in\mathcal{Q}_v$, let
\begin{equation}
\mathbf{u}^{\mathrm{self}}_{v,\tau},
\mathbf{u}^{\mathrm{multi}}_{v,\tau}
\in
\mathbb{R}^{d}
\end{equation}
denote the two retained backbone representations. A scalar gate is
computed as
\begin{equation}
g_{v,\tau}
=
\sigma
\left(
\mathbf{w}_g^\top
\left[
\mathbf{u}^{\mathrm{self}}_{v,\tau};
\mathbf{u}^{\mathrm{multi}}_{v,\tau}
\right]
+
b_g
\right),
\label{eq:app-gate}
\end{equation}
where $\mathbf{w}_g\in\mathbb{R}^{2d}$,
$b_g\in\mathbb{R}$, and $\sigma(\cdot)$ is the sigmoid function.
There is one scalar gate per query token, broadcast across all latent
features.

The fused representation is
\begin{equation}
\mathbf{u}^{\mathrm{mix}}_{v,\tau}
=
\left(
1-g_{v,\tau}
\right)
\mathbf{u}^{\mathrm{self}}_{v,\tau}
+
g_{v,\tau}
\mathbf{u}^{\mathrm{multi}}_{v,\tau}.
\label{eq:app-fused-representation}
\end{equation}
Larger gate values therefore assign greater weight to the multi-channel
representation.

A shared affine prediction head
$h_{\theta}:\mathbb{R}^{d}\rightarrow\mathbb{R}^{H}$ produces
\begin{equation}
\begin{aligned}
\widehat{\mathbf{y}}^{\mathrm{self}}_{v,\tau}
&=
h_{\theta}
\left(
\mathbf{u}^{\mathrm{self}}_{v,\tau}
\right),\\
\widehat{\mathbf{y}}^{\mathrm{multi}}_{v,\tau}
&=
h_{\theta}
\left(
\mathbf{u}^{\mathrm{multi}}_{v,\tau}
\right),\\
\widehat{\mathbf{y}}^{\mathrm{mix}}_{v,\tau}
&=
h_{\theta}
\left(
\mathbf{u}^{\mathrm{mix}}_{v,\tau}
\right).
\end{aligned}
\label{eq:app-branch-predictions}
\end{equation}

Let
\begin{equation}
\mathcal{Q}
=
\biguplus_{v\in V_{\mathrm{tok}}}
\left\{
(v,\tau):
\tau\in\mathcal{Q}_v
\right\}
\end{equation}
denote the multiset of query-token occurrences, so that
$|\mathcal{Q}|=Q$. For
$b\in\{\mathrm{self},\mathrm{multi},\mathrm{mix}\}$, the forecast loss
is
\begin{equation}
\mathcal{L}_{b}
=
\frac{1}{QH}
\sum_{(v,\tau)\in\mathcal{Q}}
\left\|
\widehat{\mathbf{y}}^{b}_{v,\tau}
-
\overline{\mathbf{y}}_{v,\tau}
\right\|_2^2.
\label{eq:app-branch-loss}
\end{equation}
No missing-value or zero-target mask is applied during synthetic
pretraining.

For numerical stability, define
\begin{equation}
g_{v,\tau}^{\epsilon}
=
\min
\left(
1-\epsilon,
\max(\epsilon,g_{v,\tau})
\right),
\qquad
\epsilon=10^{-6}.
\end{equation}
The mean binary gate entropy is
\begin{equation}
\begin{aligned}
\mathcal{H}_{\mathrm{gate}}
=
-\frac{1}{Q}
\sum_{(v,\tau)\in\mathcal{Q}}
\Big[
&
g_{v,\tau}^{\epsilon}
\log g_{v,\tau}^{\epsilon}\\
&+
\left(
1-g_{v,\tau}^{\epsilon}
\right)
\log
\left(
1-g_{v,\tau}^{\epsilon}
\right)
\Big].
\end{aligned}
\label{eq:app-gate-entropy}
\end{equation}
The pretraining objective is
\begin{equation}
\mathcal{L}
=
\mathcal{L}_{\mathrm{mix}}
+
0.3\mathcal{L}_{\mathrm{self}}
+
0.3\mathcal{L}_{\mathrm{multi}}
-
0.03\mathcal{H}_{\mathrm{gate}}.
\label{eq:app-training-objective}
\end{equation}
The auxiliary losses keep both branches individually predictive, while
the negative entropy term discourages premature gate saturation. At
inference time, only
$\widehat{\mathbf{y}}^{\mathrm{mix}}_{v,\tau}$ is returned.

\section{Additional Experimental Details}
\label{app:experimental-details}

\subsection{Datasets, Splits, and Preprocessing}
\label{app:dataset-details}

We evaluate on METR-LA and PEMS-BAY~\cite{li2018dcrnn} and on
PEMS03, PEMS04, PEMS07, and PEMS08~\cite{chen2001pems,guo2019astgcn}.
All series are treated as having nominal five-minute resolution. The
executed preprocessing retains every loaded sensor and time point and
performs no interpolation, resampling, deduplication, gap filling, or
missing-value imputation. METR-LA and PEMS-BAY use all non-timestamp columns from their CSV files.
For PEMS03, PEMS04, and PEMS08, preprocessing extracts feature index zero
from the public time-by-sensor-by-feature tensors and stores the resulting
time-by-sensor matrices. For PEMS07, the tensor representation is retained,
and feature index zero is selected during evaluation. Calendar and timestamp features are not supplied to GAS-PFN.

\begin{table*}[t]
\centering
\caption{Dataset dimensions, chronological splits, and graph inputs.
The row counts refer to raw time points. Test windows use an 18-step
forecast horizon and stride one.}
\label{tab:dataset-details}
\small
\setlength{\tabcolsep}{5pt}
\begin{tabular}{llrrcrl}
\toprule
\textbf{Dataset}
&
\textbf{Signal}
&
\textbf{Time points}
&
\textbf{Sensors}
&
\textbf{Train / val. / test rows}
&
\textbf{Test windows}
&
\textbf{Graph input}
\\
\midrule
METR-LA
& Speed
& 34,272
& 207
& 23,990 / 3,428 / 6,854
& 6,837
& Weighted, asymmetric
\\
PEMS-BAY
& Speed
& 52,116
& 325
& 36,481 / 5,212 / 10,423
& 10,406
& Weighted, asymmetric
\\
PEMS03
& Flow
& 26,208
& 358
& 18,345 / 2,622 / 5,241
& 5,224
& Binary, symmetric
\\
PEMS04
& Flow
& 16,992
& 307
& 11,894 / 1,700 / 3,398
& 3,381
& Binary, symmetric
\\
PEMS07
& Flow
& 28,224
& 883
& 19,756 / 2,824 / 5,644
& 5,627
& Weighted cost kernel, symmetric
\\
PEMS08
& Flow
& 17,856
& 170
& 12,499 / 1,786 / 3,571
& 3,554
& Binary, symmetric
\\
\bottomrule
\end{tabular}
\end{table*}

\paragraph{Chronological splits.}
For a dataset $\delta$ containing $T_{\delta}$ time points, the split
sizes are
\begin{equation}
T_{\delta}^{\mathrm{tr}}
=
\left\lfloor 0.7T_{\delta}\right\rfloor,
\qquad
T_{\delta}^{\mathrm{te}}
=
\left\lfloor 0.2T_{\delta}\right\rfloor,
\end{equation}
and
\begin{equation}
T_{\delta}^{\mathrm{val}}
=
T_{\delta}
-
T_{\delta}^{\mathrm{tr}}
-
T_{\delta}^{\mathrm{te}}.
\label{eq:app-dataset-splits}
\end{equation}
The intervals are contiguous and ordered as training, validation, and
test. Sliding windows are constructed separately for each split.
Validation and test targets remain entirely within their respective
intervals, while the first lookback in each of those intervals uses the
final 18 observations immediately preceding it.

All benchmark windows use $L=18$, $H=18$, and stride one. Reported
horizons $h\in\{6,12,18\}$ are cumulative: each result aggregates
forecast leads $1,\ldots,h$ rather than only lead $h$.

\paragraph{Normalization.}
Let $V_{\delta}$ denote the sensor set of dataset $\delta$. For each
$v\in V_{\delta}$, GAS-PFN fits a separate standardizer from the
training rows:
\begin{equation}
\mu_v^{(\delta)}
=
\frac{1}{T_{\delta}^{\mathrm{tr}}}
\sum_{t=0}^{T_{\delta}^{\mathrm{tr}}-1}
y_v^{(\delta)}(t),
\label{eq:app-dataset-mean}
\end{equation}
\begin{equation}
\left(
\sigma_v^{(\delta)}
\right)^2
=
\frac{1}{T_{\delta}^{\mathrm{tr}}}
\sum_{t=0}^{T_{\delta}^{\mathrm{tr}}-1}
\left(
y_v^{(\delta)}(t)
-
\mu_v^{(\delta)}
\right)^2,
\label{eq:app-dataset-variance}
\end{equation}
and
\begin{equation}
\widetilde{y}_v^{(\delta)}(t)
=
\frac{
y_v^{(\delta)}(t)
-
\mu_v^{(\delta)}
}{
\sigma_v^{(\delta)}
}.
\label{eq:app-dataset-standardization}
\end{equation}
The same sensor-specific transformation is applied to training supports
and test queries. Predictions and targets are inverse-transformed before
metric computation. The synthetic task-level robust scaling in
Eq.~\eqref{eq:app-task-normalization} is not applied to benchmark data,
and no output clipping is performed.

\paragraph{Dataset graph inputs.}
GAS-PFN consumes the stored adjacency matrix associated with each
benchmark. Its diagonal is set to one before neighborhood routing; the
matrix is not otherwise symmetrized, binarized, thresholded, or
degree-normalized.

The METR-LA and PEMS-BAY matrices are weighted and numerically
asymmetric. PEMS03, PEMS04, and PEMS08 use symmetric binary
connectivity; available edge attributes beyond connectivity are not
used. For PEMS07, an undirected edge with positive recorded cost
$\operatorname{cost}(v,v')$ is assigned
\begin{equation}
A_{vv'}
=
\exp
\left(
-\frac{
\operatorname{cost}(v,v')
}{
\sigma_{\mathrm{cost}}
}
\right),
\label{eq:app-pems07-adjacency}
\end{equation}
where $\sigma_{\mathrm{cost}}$ is the standard deviation of the retained
positive edge costs. Parallel entries retain their maximum weight, and
no threshold is applied. These stored graph matrices are distinct from
the row-scaled graph scores in
Eq.~\eqref{eq:app-row-scaled-graph}.

\subsection{Evaluation Metrics}
\label{app:metric-details}

Let $W_{\delta}^{\mathrm{te}}$ denote the number of test windows for
dataset $\delta$. Let
$\widehat{y}^{(\delta)}_{u,s,v}$ and
$y^{(\delta)}_{u,s,v}$ denote the original-scale prediction and ground
truth for test window
$u\in\{1,\ldots,W_{\delta}^{\mathrm{te}}\}$, forecast lead
$s\in\{1,\ldots,H\}$, and sensor $v\in V_{\delta}$.

For cumulative horizon $h$, define
\begin{equation}
\Omega_{\delta}(h)
=
\left\{
(u,s,v):
\begin{array}{l}
1\leq u\leq W_{\delta}^{\mathrm{te}},\\
1\leq s\leq h,\\
v\in V_{\delta},\\
y^{(\delta)}_{u,s,v}\neq 0
\end{array}
\right\}.
\label{eq:app-valid-metric-set}
\end{equation}
The reported metrics are
\begin{equation}
\operatorname{MAE}_{\delta}(h)
=
\frac{1}{|\Omega_{\delta}(h)|}
\sum_{(u,s,v)\in\Omega_{\delta}(h)}
\left|
\widehat{y}^{(\delta)}_{u,s,v}
-
y^{(\delta)}_{u,s,v}
\right|,
\label{eq:app-masked-mae}
\end{equation}
and
\begin{equation}
\operatorname{MSE}_{\delta}(h)
=
\frac{1}{|\Omega_{\delta}(h)|}
\sum_{(u,s,v)\in\Omega_{\delta}(h)}
\left(
\widehat{y}^{(\delta)}_{u,s,v}
-
y^{(\delta)}_{u,s,v}
\right)^2.
\label{eq:app-masked-mse}
\end{equation}
The zero comparison is exact and is applied to raw ground truth after
inverse transformation. For each dataset, MAE and MSE are computed by pooling errors across all
valid test windows, forecast steps $1,\ldots,h$, and sensors. Cross-dataset
averages are unweighted arithmetic means of the six dataset-level
metrics.

Zero masking in Eqs.~\eqref{eq:app-masked-mae} and
\eqref{eq:app-masked-mse} applies to benchmark evaluation. GAS-PFN's
synthetic pretraining objective does not use a benchmark zero-value
mask.

\subsection{GAS-PFN Configuration and Evaluation}
\label{app:gaspfn-experimental-setup}

Hyperparameters governing synthetic-prior generation are reported in
Appendix~\ref{app:synthetic-prior}.
Table~\ref{tab:gaspfn-hyperparameters} summarizes the architecture,
optimization, and target-evaluation settings.

\begin{table*}[t]
\centering
\caption{GAS-PFN architecture, pretraining, and evaluation
hyperparameters.}
\label{tab:gaspfn-hyperparameters}
\small
\setlength{\tabcolsep}{7pt}
\begin{tabular}{lll}
\toprule
\textbf{Category} & \textbf{Setting} & \textbf{Value} \\
\midrule
Architecture
& Lookback / forecast length
& $18/18$
\\
Architecture
& Transformer blocks / attention heads
& $10/8$
\\
Architecture
& Hidden / feed-forward dimension
& $512/2048$
\\
Architecture
& Attention-head dimension
& $64$
\\
Architecture
& Dropout
& $0.1$
\\
Architecture
& Temporal-distance initialization
& $\gamma_h=0.01$
\\
Architecture
& Router mixture $(1-\eta)/\eta$
& $0.6/0.4$
\\
Architecture
& Routed neighborhood
& Up to $7$ selected channels
\\
Architecture
& Fusion
& Scalar gate per query token
\\
Architecture
& Model parameters
& $32{,}100{,}395$
\\
\midrule
Pretraining
& Context cap per represented sensor
& $8$ recent raw $+4$ older raw $+4$ pooled memory
\\
Pretraining
& Optimizer
& AdamW
\\
Pretraining
& Learning rate / weight decay
& $10^{-4}/0.05$
\\
Pretraining
& AdamW $(\beta_1,\beta_2,\epsilon)$
& $(0.9,0.95,10^{-8})$
\\
Pretraining
& Learning-rate schedule
& $50{,}000$-step warmup, then cosine decay to $10^{-5}$
\\
Pretraining
& Meta-task batch / gradient accumulation
& $1$/none
\\
Pretraining
& Gradient clipping
& Global norm $1.0$
\\
Pretraining
& Training length
& $1{,}000{,}000$ steps
  ($400$ epochs $\times$ $2{,}500$ tasks)
\\
Pretraining
& Precision
& Float32
\\
Pretraining
& Reported checkpoint
& Terminal state at step $1{,}000{,}000$ (epoch $400$)
\\
\midrule
Evaluation
& Target-domain support per sensor
& $16$ recent raw $+16$ historical raw patches
\\
Evaluation
& Query batch size
& $4$ consecutive forecast origins
\\
Evaluation
& Multi-branch query attention
& Same forecast origin only
\\
\bottomrule
\end{tabular}
\end{table*}

The three primary GAS-PFN checkpoints are independently pretrained with
seeds 42, 43, and 44. Each checkpoint is the terminal state after one
million optimizer steps. The frozen synthetic bank is shared across
runs, while task sampling and optimization are independently seeded.

\paragraph{Target-domain context.}
For each sensor $v$, the main zero-shot evaluation uses the retained
endpoint set $\widetilde{\mathcal{C}}_v$ defined in
Eq.~\eqref{eq:app-target-context-set}: 16 aligned recent supports and 16
sensor-specific historical anchors. The context is reconstructed
identically for each checkpoint and cached for every test query. No
attention-pooled memory tokens are used during benchmark evaluation.

\paragraph{Full-channel and PEMS07 inference.}
METR-LA, PEMS-BAY, PEMS03, PEMS04, and PEMS08 are evaluated with all
sensors present in each forward pass. PEMS07 uses memory-bounded channel-cohort inference, while the other
five datasets use full-channel inference. For the channel cohort strategy, target sensors are grouped
in sets of at most 50, and each group's active context pool contains the
union of the targets' selected one-hop channel neighborhoods. The groups are assigned via a greedy algorithm designed to minimize repeated processing of the same channels for context. Every active sensor retains all 32 raw context patches.

This fallback preserves the complete selected context and planned
context neighborhood of each target sensor. Query tokens are retained only
for the cohort's target sensors, and representations and routing are
recomputed within the reduced token universe. The reported PEMS07
results therefore use cohort inference; the other five datasets use
full-channel inference.

\paragraph{Checkpoint aggregation.}
Let $a_r$ denote a metric obtained from checkpoint seed
$r\in\{42,43,44\}$. The primary GAS-PFN table reports
\begin{equation}
\begin{aligned}
\overline{a}
&=
\frac{1}{3}
\sum_{r\in\{42,43,44\}}
a_r, \\
\sigma_{\mathrm{pop}}
&=
\left[
\frac{1}{3}
\sum_{r\in\{42,43,44\}}
\left(
a_r-\overline{a}
\right)^2
\right]^{1/2}.
\end{aligned}
\label{eq:app-checkpoint-aggregation}
\end{equation}

\subsection{Zero-Shot Baseline Configuration}
\label{app:baseline-configuration}

Chronos-2~\cite{ansari2025chronos2},
TimesFM-2.5~\cite{das2024timesfm},
Moirai-2~\cite{liu2025moirai2},
Lag-Llama~\cite{rasul2024lagllama}, and
TimePFN~\cite{taga2025timepfn} are each evaluated under a recent-only
and a history-augmented context regime. Except for model-specific input
requirements described below, the recent-only regime uses the 18
observations immediately preceding each forecast origin. The
history-augmented regime concatenates a contiguous prefix from the end
of the 70\% training split with the same 18 recent observations. The
training prefix and recent window are disjoint.

For each baseline and dataset, both regimes are rescored using
Eqs.~\eqref{eq:app-masked-mae} and \eqref{eq:app-masked-mse}. We report
the regime with lower cumulative 18-step test MAE and use that single
regime for every reported horizon and both metrics. The lower-MAE regime is selected once per baseline and dataset and is
then used for every reported horizon and both metrics. GAS-PFN uses one
fixed 32-patch support configuration.

\begin{table*}[t]
\centering
\caption{External zero-shot baseline implementations and context
regimes. ``Training tail'' denotes observations drawn contiguously from
the end of the 70\% training split. For Lag-Llama, the raw-history totals
include the additional buffer required to construct its fixed lag
features.}
\label{tab:baseline-contexts}
\footnotesize
\setlength{\tabcolsep}{5pt}
\begin{tabular}{
p{0.14\textwidth}
p{0.29\textwidth}
p{0.22\textwidth}
p{0.25\textwidth}}
\toprule
\textbf{Model}
&
\textbf{Implementation}
&
\textbf{Recent-only regime}
&
\textbf{History-augmented regime}
\\
\midrule

Chronos-2
&
\texttt{amazon/chronos-2};
\texttt{chronos-forecasting 2.2.2}
&
18 recent observations
&
8,174 training-tail $+18$ recent $=8,192$
\\

TimesFM-2.5
&
\texttt{google/timesfm-2.5-}\newline
\texttt{200m-pytorch};\newline
\texttt{timesfm 2.0.0}
&
18 recent observations
&
78 training-tail $+18$ recent $=96$
\\

Moirai-2
&
\texttt{Salesforce/moirai-2.0-}\newline
\texttt{R-small};\newline
\texttt{uni2ts 2.0.0}
&
18 recent observations
&
1,662 training-tail $+18$ recent $=1,680$
\\

Lag-Llama
&
Released \texttt{lag-llama.ckpt}
&
Nominal context: 18 recent;
1,110 raw observations including lag buffer
&
Nominal context: 14 training-tail $+18$ recent $=32$;
1,124 raw observations including lag buffer
\\

TimePFN
&
Released TimePFN checkpoint with benchmark adapters
&
18 recent observations, left-padded to length 96
&
78 training-tail $+18$ recent $=96$
\\

\bottomrule
\end{tabular}
\end{table*}

Chronos-2 is evaluated through its multivariate data-frame interface
without exogenous covariates. TimesFM-2.5 is applied independently to
each sensor with its native input normalization and positive-output
configuration. Moirai-2 is also applied independently by sensor,
without static or dynamic covariates, and its median quantile is used as
the point forecast. TimePFN is evaluated as a multivariate model over
contiguous sensor blocks of at most 161 channels, with predictions
restored to the original sensor order. All external prediction arrays
are inverse-transformed by their corresponding wrappers and rescored
using the common raw-scale metric implementation.

The external pretrained checkpoints use deterministic inference paths
and are therefore reported as single values. Chronos-2, TimesFM-2.5,
and Moirai-2 use their $0.5$-quantile forecasts, Lag-Llama uses its
analytic predictive mean, and TimePFN uses its direct point output;
repeated executions produced byte-identical predictions.

\paragraph{Lag-Llama context handling.}
Lag-Llama constructs its inputs using a fixed set of lagged values.
Accordingly, we preserve the released model's native lag-feature
construction by providing the additional historical buffer required to
instantiate those lags. Its recent-only and history-augmented
configurations therefore consume 1,110 and 1,124 raw observations,
respectively, even though their designated model-context lengths are 18
and 32 and the contiguous recent window remains 18 observations. This preserves the released model's native lag-feature construction
without modifying its architecture.

\subsection{Additional Comparison Protocols}
\label{app:additional-comparison-protocols}

\paragraph{TimePFN fine-tuning.}
The TimePFN fine-tuning comparison uses labeled-window budgets of 32 and
64 on METR-LA, PEMS-BAY, PEMS03, PEMS04, and PEMS08; PEMS07 is excluded
because the registered fine-tuning configuration exceeded available
GPU memory. Budgets are nested, and the same sampled window indices are
shared across the two TimePFN context variants for each seed.

All TimePFN parameters are updated using AdamW with learning rate
$2\times10^{-4}$, weight decay $0.01$, batch size 16, and at most eight
epochs. The learning rate is halved after each epoch, early-stopping
patience is three epochs, and checkpoint selection uses validation
performance. Training and validation minimize original-scale,
exact-zero-masked MAE. Results are aggregated over seeds 42, 43, and 44.

\paragraph{SCPT selected-road evaluation.}
The cross-road experiment uses SCPT~\cite{prabowo2023scpt} with a
12-step lookback and an 18-step output. For each dataset, the fraction of roads
available for SCPT training ranges from 10\% to 80\%; 10\% of roads are
reserved for validation, and the remaining roads form the unseen test
set. Seeds 42, 43, and 44 define independent road permutations. Each
GAS-PFN result is paired with the checkpoint having the same seed, while
SCPT is trained separately for that road partition.

SCPT retains its native preprocessing, which computes one scalar mean
and standard deviation over the complete time-by-road matrix before the
temporal and road splits. GAS-PFN instead uses the train-only
sensor-specific scaling in Eqs.~\eqref{eq:app-dataset-mean}--
\eqref{eq:app-dataset-standardization}. For the final comparison, both methods are
inverse-transformed, aligned to the same raw target tensor, and rescored
using the identical exact-zero mask.

The selected-road results report the arithmetic mean and sample standard
deviation over the three paired model and road-partition seeds. This
variation therefore includes both independently pretrained GAS-PFN
checkpoints and different supervised-road partitions.

\section{Additional Benchmark Results}

\subsection{Additional Zero-shot Results}

\label{app:additional-results}

Table~\ref{tab:zs_context_comparison_306090} reports both evaluated context regimes for every external baseline across all datasets and forecast horizons. ``Recent 18'' uses a contiguous recent window of 18 observations, while ``Recent + prefix'' additionally supplies a contiguous tail from the training split. The Total column reports raw temporal input per channel. For Lag-Llama, marked by $\dagger$, this total also includes the longer raw-history buffer required by its native lag features, as detailed in Appendix~\ref{app:experimental-details}. GAS-PFN's total counts 18 recent values together with 32 support patches of 36 scalar values each, yielding 1,170 scalar input slots.

These expanded results reinforce the main-paper observation that more contiguous history is not uniformly beneficial: the effect of adding a training-tail prefix varies substantially across models and datasets. GAS-PFN instead uses one fixed graph-aware mixed-context construction, supporting the claim that the organization and use of historical context matter in addition to its raw quantity.

\paragraph{Computing infrastructure.}
All experiments were run as single-GPU jobs on a server equipped with
eight NVIDIA A100-SXM4-80GB GPUs, AMD EPYC 7742 processors, and
approximately 1.97 TiB of system memory. The server ran Ubuntu 24.04.4
LTS. GAS-PFN training and evaluation used CPython 3.12.3 and PyTorch
2.6.0+cu124 with CUDA 12.4.

\begin{table*}[!t]
\centering

\caption{
Zero-shot MAE and MSE under recent-only and history-augmented context regimes at cumulative 30-, 60-, and 90-minute horizons. Lower is better. Bold and underlined values denote the best and second-best results, respectively, using unrounded values.
}
\label{tab:zs_context_comparison_306090}

\begingroup
\scriptsize
\setlength{\tabcolsep}{2.2pt}
\renewcommand{\arraystretch}{1.12}

\begin{tabular}{@{}lllcccccc@{}}
\toprule
\multirow{2}{*}{\textbf{Model}}
& \multirow{2}{*}{\textbf{Context}}
& \multirow{2}{*}{\textbf{Total}}
& \multicolumn{2}{c}{\textbf{30 min}}
& \multicolumn{2}{c}{\textbf{60 min}}
& \multicolumn{2}{c}{\textbf{90 min}} \\
\cmidrule(lr){4-5}
\cmidrule(lr){6-7}
\cmidrule(lr){8-9}
& &
& \textbf{MAE} & \textbf{MSE}
& \textbf{MAE} & \textbf{MSE}
& \textbf{MAE} & \textbf{MSE} \\
\midrule

% =========================================================
% METR-LA
% =========================================================
\multicolumn{9}{l}{\textbf{METR-LA}} \\*
\addlinespace[1pt]

\textbf{GAS-PFN}
& Mixed
& 1,170
& $3.97 \pm 0.07$
& $\mathbf{68.31} \pm 2.35$
& $4.83 \pm 0.07$
& $97.61 \pm 3.58$
& $5.50 \pm 0.06$
& $119.69 \pm 5.08$ \\*

\addlinespace[2pt]

\multirow{2}{*}{Chronos-2}
& Recent 18
& 18
& 4.31 & 91.60
& 5.74 & 152.61
& 6.66 & 193.68 \\
& Recent + prefix
& 8,192
& \textbf{3.68} & 70.62
& \textbf{4.38} & \textbf{91.23}
& \textbf{4.95} & \textbf{107.42} \\*

\addlinespace[1pt]

\multirow{2}{*}{TimesFM-2.5}
& Recent 18
& 18
& 4.10 & 86.15
& 5.19 & 133.05
& 6.06 & 171.30 \\
& Recent + prefix
& 96
& \underline{3.80} & 72.30
& \underline{4.73} & 108.38
& \underline{5.45} & 136.04 \\*

\addlinespace[1pt]

\multirow{2}{*}{Moirai-2}
& Recent 18
& 18
& 4.23 & 83.92
& 5.30 & 124.56
& 6.35 & 163.35 \\
& Recent + prefix
& 1,680
& 4.25 & 77.87
& 5.63 & 123.07
& 6.53 & 154.96 \\*

\addlinespace[1pt]

\multirow{2}{*}{Lag-Llama}
& Recent 18$^\dagger$
& 1,110
& 4.86 & 86.23
& 6.21 & 130.31
& 7.41 & 178.32 \\
& Recent + prefix$^\dagger$
& 1,124
& 6.69 & 230.75
& 7.42 & 255.05
& 8.02 & 277.87 \\*

\addlinespace[1pt]

\multirow{2}{*}{TimePFN}
& Recent 18
& 18
& 5.84 & 77.06
& 6.59 & 102.16
& 7.16 & 118.44 \\
& Recent + prefix
& 96
& 4.86 & \underline{69.41}
& 5.59 & \underline{94.33}
& 6.13 & \underline{112.73} \\*

\midrule

% =========================================================
% PEMS-BAY
% =========================================================
\multicolumn{9}{l}{\textbf{PEMS-BAY}} \\*
\addlinespace[1pt]

\textbf{GAS-PFN}
& Mixed
& 1,170
& $1.73 \pm 0.01$
& $\underline{14.85} \pm 0.12$
& $\underline{2.21} \pm 0.02$
& $\mathbf{25.22} \pm 0.11$
& $\mathbf{2.59} \pm 0.01$
& $\mathbf{34.45} \pm 0.09$ \\*

\addlinespace[2pt]

\multirow{2}{*}{Chronos-2}
& Recent 18
& 18
& 1.77 & 16.28
& 2.50 & 35.14
& 2.98 & 49.30 \\
& Recent + prefix
& 8,192
& \textbf{1.61} & \textbf{12.90}
& 2.24 & \underline{27.25}
& 2.71 & 40.58 \\*

\addlinespace[1pt]

\multirow{2}{*}{TimesFM-2.5}
& Recent 18
& 18
& 1.78 & 18.20
& 2.40 & 34.61
& 2.88 & 49.03 \\
& Recent + prefix
& 96
& \underline{1.65} & 15.01
& \textbf{2.18} & 27.70
& \underline{2.62} & \underline{39.32} \\*

\addlinespace[1pt]

\multirow{2}{*}{Moirai-2}
& Recent 18
& 18
& 1.74 & 15.53
& 2.36 & 29.99
& 2.91 & 44.67 \\
& Recent + prefix
& 1,680
& 1.85 & 17.52
& 2.66 & 40.00
& 3.17 & 55.69 \\*

\addlinespace[1pt]

\multirow{2}{*}{Lag-Llama}
& Recent 18$^\dagger$
& 1,110
& 2.58 & 41.38
& 3.01 & 48.85
& 3.35 & 55.81 \\
& Recent + prefix$^\dagger$
& 1,124
& 2.59 & 45.83
& 2.99 & 52.64
& 3.29 & 58.78 \\*

\addlinespace[1pt]

\multirow{2}{*}{TimePFN}
& Recent 18
& 18
& 2.62 & 22.67
& 3.03 & 32.66
& 3.37 & 40.82 \\
& Recent + prefix
& 96
& 2.41 & 22.35
& 2.85 & 32.51
& 3.21 & 41.51 \\*

\midrule

% =========================================================
% PEMS03
% =========================================================
\multicolumn{9}{l}{\textbf{PEMS03}} \\*
\addlinespace[1pt]

\textbf{GAS-PFN}
& Mixed
& 1,170
& $\mathbf{17.57} \pm 0.60$
& $\underline{749.58} \pm 36.52$
& $\mathbf{21.51} \pm 0.92$
& $\mathbf{1160.61} \pm 69.29$
& $\mathbf{25.97} \pm 1.35$
& $\mathbf{1740.93} \pm 129.80$ \\*

\addlinespace[2pt]

\multirow{2}{*}{Chronos-2}
& Recent 18
& 18
& 19.11 & 923.53
& 25.46 & 1696.22
& 31.02 & 2510.18 \\
& Recent + prefix
& 8,192
& \underline{17.74} & \textbf{747.80}
& \underline{23.22} & \underline{1320.70}
& \underline{28.98} & \underline{2080.14} \\*

\addlinespace[1pt]

\multirow{2}{*}{TimesFM-2.5}
& Recent 18
& 18
& 19.54 & 937.01
& 25.82 & 1699.79
& 31.91 & 2638.02 \\
& Recent + prefix
& 96
& 19.52 & 894.14
& 25.26 & 1496.84
& 31.61 & 2357.27 \\*

\addlinespace[1pt]

\multirow{2}{*}{Moirai-2}
& Recent 18
& 18
& 19.64 & 913.24
& 25.74 & 1646.29
& 32.35 & 2666.41 \\
& Recent + prefix
& 1,680
& 22.44 & 1135.31
& 32.96 & 2457.30
& 41.67 & 3812.48 \\*

\addlinespace[1pt]

\multirow{2}{*}{Lag-Llama}
& Recent 18$^\dagger$
& 1,110
& 20.79 & 993.79
& 29.50 & 2006.43
& 39.31 & 3556.25 \\
& Recent + prefix$^\dagger$
& 1,124
& 20.23 & 918.27
& 27.25 & 1672.31
& 34.66 & 2709.34 \\*

\addlinespace[1pt]

\multirow{2}{*}{TimePFN}
& Recent 18
& 18
& 34.50 & 2201.97
& 38.46 & 2788.31
& 42.60 & 3417.89 \\
& Recent + prefix
& 96
& 39.47 & 2742.86
& 43.29 & 3366.56
& 48.19 & 4240.31 \\*

\bottomrule
\end{tabular}
\endgroup
\end{table*}

% =====================================================================
% CONTINUED TABLE
% =====================================================================

\begin{table*}[!t]
\centering

{\scriptsize\itshape
Table~\ref{tab:zs_context_comparison_306090} continued
\par}
\vspace{2pt}

\begingroup
\scriptsize
\setlength{\tabcolsep}{2.2pt}
\renewcommand{\arraystretch}{1.12}

\begin{tabular}{@{}lllcccccc@{}}
\toprule
\multirow{2}{*}{\textbf{Model}}
& \multirow{2}{*}{\textbf{Context}}
& \multirow{2}{*}{\textbf{Total}}
& \multicolumn{2}{c}{\textbf{30 min}}
& \multicolumn{2}{c}{\textbf{60 min}}
& \multicolumn{2}{c}{\textbf{90 min}} \\
\cmidrule(lr){4-5}
\cmidrule(lr){6-7}
\cmidrule(lr){8-9}
& &
& \textbf{MAE} & \textbf{MSE}
& \textbf{MAE} & \textbf{MSE}
& \textbf{MAE} & \textbf{MSE} \\
\midrule

% =========================================================
% PEMS04
% =========================================================
\multicolumn{9}{l}{\textbf{PEMS04}} \\*
\addlinespace[1pt]

\textbf{GAS-PFN}
& Mixed
& 1,170
& $\mathbf{22.52} \pm 0.28$
& $\mathbf{1243.43} \pm 13.35$
& $\mathbf{27.16} \pm 0.41$
& $\mathbf{1782.98} \pm 26.45$
& $\mathbf{32.32} \pm 0.53$
& $\mathbf{2517.88} \pm 54.69$ \\*

\addlinespace[2pt]

\multirow{2}{*}{Chronos-2}
& Recent 18
& 18
& \underline{24.23} & 1402.54
& \underline{31.26} & 2300.66
& \underline{37.92} & \underline{3370.61} \\
& Recent + prefix
& 8,192
& 25.61 & \underline{1402.39}
& 32.89 & \underline{2266.07}
& 40.67 & 3466.71 \\*

\addlinespace[1pt]

\multirow{2}{*}{TimesFM-2.5}
& Recent 18
& 18
& 25.33 & 1491.79
& 32.81 & 2476.17
& 40.18 & 3712.61 \\
& Recent + prefix
& 96
& 25.25 & 1586.96
& 31.77 & 2514.64
& 38.58 & 3708.20 \\*

\addlinespace[1pt]

\multirow{2}{*}{Moirai-2}
& Recent 18
& 18
& 25.90 & 1518.06
& 33.45 & 2523.53
& 41.73 & 3966.89 \\
& Recent + prefix
& 1,680
& 28.88 & 1734.25
& 40.59 & 3390.69
& 52.15 & 5730.72 \\*

\addlinespace[1pt]

\multirow{2}{*}{Lag-Llama}
& Recent 18$^\dagger$
& 1,110
& 27.62 & 1698.79
& 38.54 & 3152.43
& 50.72 & 5371.03 \\
& Recent + prefix$^\dagger$
& 1,124
& 33.65 & 2411.36
& 54.64 & 6231.24
& 74.20 & 11249.26 \\*

\addlinespace[1pt]

\multirow{2}{*}{TimePFN}
& Recent 18
& 18
& 43.19 & 3218.50
& 47.91 & 3987.86
& 52.92 & 4841.89 \\
& Recent + prefix
& 96
& 61.26 & 7452.77
& 66.46 & 8623.08
& 73.41 & 10500.47 \\*

\midrule

% =========================================================
% PEMS07
% =========================================================
\multicolumn{9}{l}{\textbf{PEMS07}} \\*
\addlinespace[1pt]

\textbf{GAS-PFN}
& Mixed
& 1,170
& $\mathbf{25.15} \pm 0.54$
& $\mathbf{1493.64} \pm 33.87$
& $\mathbf{30.81} \pm 0.75$
& $\mathbf{2223.93} \pm 56.13$
& $\mathbf{37.15} \pm 0.98$
& $\mathbf{3239.16} \pm 94.62$ \\*

\addlinespace[2pt]

\multirow{2}{*}{Chronos-2}
& Recent 18
& 18
& 26.69 & 1726.37
& 35.12 & 3000.69
& 42.82 & 4450.07 \\
& Recent + prefix
& 8,192
& \underline{25.95} & \underline{1522.68}
& \underline{32.76} & \underline{2366.80}
& \underline{40.01} & \underline{3558.66} \\*

\addlinespace[1pt]

\multirow{2}{*}{TimesFM-2.5}
& Recent 18
& 18
& 27.95 & 1838.07
& 37.16 & 3274.80
& 46.14 & 5088.31 \\
& Recent + prefix
& 96
& 26.25 & 1612.32
& 33.96 & 2627.48
& 42.16 & 4053.21 \\*

\addlinespace[1pt]

\multirow{2}{*}{Moirai-2}
& Recent 18
& 18
& 28.74 & 1858.14
& 38.36 & 3342.03
& 48.91 & 5490.79 \\
& Recent + prefix
& 1,680
& 31.90 & 2371.38
& 47.45 & 5832.52
& 61.01 & 9626.35 \\*

\addlinespace[1pt]

\multirow{2}{*}{Lag-Llama}
& Recent 18$^\dagger$
& 1,110
& 31.03 & 2071.72
& 44.76 & 4168.11
& 59.52 & 7317.76 \\
& Recent + prefix$^\dagger$
& 1,124
& 29.20 & 1911.20
& 39.99 & 3328.94
& 50.77 & 5272.50 \\*

\addlinespace[1pt]

\multirow{2}{*}{TimePFN}
& Recent 18
& 18
& 50.69 & 4168.78
& 56.65 & 5309.43
& 62.77 & 6544.13 \\
& Recent + prefix
& 96
& 46.42 & 4035.22
& 52.46 & 5170.30
& 58.75 & 6523.84 \\*

\midrule

% =========================================================
% PEMS08
% =========================================================
\multicolumn{9}{l}{\textbf{PEMS08}} \\*
\addlinespace[1pt]

\textbf{GAS-PFN}
& Mixed
& 1,170
& $\mathbf{17.78} \pm 0.19$
& $\mathbf{767.87} \pm 17.31$
& $\mathbf{21.35} \pm 0.36$
& $\mathbf{1112.37} \pm 41.18$
& $\mathbf{25.21} \pm 0.47$
& $\mathbf{1548.52} \pm 71.32$ \\*

\addlinespace[2pt]

\multirow{2}{*}{Chronos-2}
& Recent 18
& 18
& 19.66 & 938.59
& 25.87 & 1634.48
& 31.55 & 2393.38 \\
& Recent + prefix
& 8,192
& 19.10 & 834.07
& 24.15 & \underline{1305.08}
& 29.44 & \underline{1922.89} \\*

\addlinespace[1pt]

\multirow{2}{*}{TimesFM-2.5}
& Recent 18
& 18
& 20.41 & 983.52
& 26.94 & 1715.18
& 33.37 & 2620.24 \\
& Recent + prefix
& 96
& \underline{18.50} & \underline{816.30}
& \underline{23.54} & 1313.15
& \underline{28.60} & 1924.25 \\*

\addlinespace[1pt]

\multirow{2}{*}{Moirai-2}
& Recent 18
& 18
& 20.44 & 961.84
& 26.77 & 1666.28
& 33.96 & 2695.22 \\
& Recent + prefix
& 1,680
& 24.60 & 1400.30
& 35.85 & 3230.58
& 45.13 & 5152.02 \\*

\addlinespace[1pt]

\multirow{2}{*}{Lag-Llama}
& Recent 18$^\dagger$
& 1,110
& 22.67 & 1121.92
& 32.68 & 2297.85
& 43.51 & 4055.36 \\
& Recent + prefix$^\dagger$
& 1,124
& 26.21 & 1392.03
& 40.16 & 3152.78
& 53.65 & 5579.32 \\*

\addlinespace[1pt]

\multirow{2}{*}{TimePFN}
& Recent 18
& 18
& 36.59 & 2299.04
& 40.71 & 2861.44
& 45.09 & 3500.04 \\
& Recent + prefix
& 96
& 26.94 & 1488.01
& 31.22 & 2031.89
& 36.54 & 2864.76 \\*

\bottomrule
\end{tabular}
\endgroup
\end{table*}

\subsection{TimePFN Fine-Tuning Comparison}

Table~\ref{tab:timepfn_finetune_comparison} compares zero-shot GAS-PFN with TimePFN after target-specific fine-tuning using 32 or 64 labeled patches. For TimePFN, ``Recent 18'' uses only the 18 observations immediately preceding each forecast origin, while ``Recent + prefix'' also supplies a contiguous tail from the training split. GAS-PFN instead uses its fixed mixed-context configuration and performs no target-specific parameter updates. Across the five evaluated datasets, GAS-PFN achieves the best mean result in 27 of the 30 dataset--horizon--metric comparisons. Fine-tuning TimePFN with 64 patches narrows the gap and yields the best 90-minute MSE on PEMS03, PEMS04, and PEMS08, but does not otherwise surpass GAS-PFN. These results reinforce that the graph-aware prior and in-context architecture remain competitive even against limited target-domain parameter adaptation.

\begin{table*}[!t]
\centering

\caption{
Comparison of zero-shot GAS-PFN with TimePFN fine-tuned using 32 or 64 target-domain patches. Results are mean $\pm$ sample standard deviation over seeds 42, 43, and 44. Lower is better; bold and underlined values denote the best and second-best means, respectively, using unrounded results.
}
\label{tab:timepfn_finetune_comparison}

\begingroup
\scriptsize
\setlength{\tabcolsep}{0.75pt}
\renewcommand{\arraystretch}{1.04}

\begin{tabular}{@{}lllcccccc@{}}
\toprule
\multirow{2}{*}{\textbf{Model}}
& \multirow{2}{*}{\textbf{Budget}}
& \multirow{2}{*}{\textbf{Context}}
& \multicolumn{2}{c}{\textbf{30 min}}
& \multicolumn{2}{c}{\textbf{60 min}}
& \multicolumn{2}{c}{\textbf{90 min}} \\
\cmidrule(lr){4-5}
\cmidrule(lr){6-7}
\cmidrule(lr){8-9}
& &
& \textbf{MAE} & \textbf{MSE}
& \textbf{MAE} & \textbf{MSE}
& \textbf{MAE} & \textbf{MSE} \\
\midrule

% =========================================================
% METR-LA
% =========================================================
\multicolumn{9}{l}{\textbf{METR-LA}} \\*

\textbf{GAS-PFN}
& --
& Mixed
& $\mathbf{3.97} \pm 0.07$
& $\mathbf{68.31} \pm 2.35$
& $\mathbf{4.83} \pm 0.07$
& $\mathbf{97.61} \pm 3.58$
& $\mathbf{5.50} \pm 0.06$
& $\mathbf{119.69} \pm 5.08$ \\*

\addlinespace[1pt]

\multirow{4}{*}{TimePFN}
& \multirow{2}{*}{FT-32}
& Recent 18
& $5.12 \pm 0.67$
& $112.67 \pm 24.70$
& $6.07 \pm 0.62$
& $155.51 \pm 26.45$
& $6.78 \pm 0.56$
& $186.47 \pm 29.54$ \\

&
& Recent + prefix
& $5.14 \pm 0.50$
& $107.27 \pm 16.52$
& $6.06 \pm 0.48$
& $149.42 \pm 19.17$
& $6.76 \pm 0.43$
& $181.29 \pm 21.37$ \\

& \multirow{2}{*}{FT-64}
& Recent 18
& $4.68 \pm 0.21$
& $97.01 \pm 7.15$
& $5.58 \pm 0.16$
& $134.08 \pm 10.82$
& $6.33 \pm 0.13$
& $163.68 \pm 14.66$ \\

&
& Recent + prefix
& $\underline{4.64} \pm 0.10$
& $\underline{91.68} \pm 3.27$
& $\underline{5.57} \pm 0.08$
& $\underline{127.98} \pm 4.02$
& $\underline{6.31} \pm 0.07$
& $\underline{157.84} \pm 5.35$ \\*

\midrule

% =========================================================
% PEMS-BAY
% =========================================================
\multicolumn{9}{l}{\textbf{PEMS-BAY}} \\*

\textbf{GAS-PFN}
& --
& Mixed
& $\mathbf{1.73} \pm 0.01$
& $\mathbf{14.85} \pm 0.12$
& $\mathbf{2.21} \pm 0.02$
& $\mathbf{25.22} \pm 0.11$
& $\mathbf{2.59} \pm 0.01$
& $\mathbf{34.45} \pm 0.09$ \\*

\addlinespace[1pt]

\multirow{4}{*}{TimePFN}
& \multirow{2}{*}{FT-32}
& Recent 18
& $1.89 \pm 0.02$
& $17.55 \pm 0.52$
& $2.34 \pm 0.04$
& $28.70 \pm 1.29$
& $2.71 \pm 0.04$
& $38.29 \pm 2.05$ \\

&
& Recent + prefix
& $1.93 \pm 0.03$
& $17.44 \pm 0.49$
& $2.39 \pm 0.05$
& $28.48 \pm 1.42$
& $2.78 \pm 0.05$
& $38.23 \pm 1.90$ \\

& \multirow{2}{*}{FT-64}
& Recent 18
& $\underline{1.79} \pm 0.01$
& $15.87 \pm 0.05$
& $\underline{2.24} \pm 0.03$
& $\underline{26.15} \pm 0.15$
& $\underline{2.61} \pm 0.04$
& $\underline{35.30} \pm 0.46$ \\

&
& Recent + prefix
& $1.82 \pm 0.02$
& $\underline{15.83} \pm 0.21$
& $2.28 \pm 0.02$
& $26.20 \pm 0.12$
& $2.66 \pm 0.02$
& $35.59 \pm 0.16$ \\*

\midrule

% =========================================================
% PEMS03
% =========================================================
\multicolumn{9}{l}{\textbf{PEMS03}} \\*

\textbf{GAS-PFN}
& --
& Mixed
& $\mathbf{17.57} \pm 0.60$
& $\mathbf{749.58} \pm 36.52$
& $\mathbf{21.51} \pm 0.92$
& $\mathbf{1160.61} \pm 69.29$
& $\mathbf{25.97} \pm 1.35$
& $1740.93 \pm 129.80$ \\*

\addlinespace[1pt]

\multirow{4}{*}{TimePFN}
& \multirow{2}{*}{FT-32}
& Recent 18
& $21.09 \pm 0.20$
& $1094.25 \pm 25.23$
& $25.70 \pm 0.40$
& $1677.81 \pm 69.23$
& $30.19 \pm 0.43$
& $2298.73 \pm 101.94$ \\

&
& Recent + prefix
& $22.38 \pm 0.45$
& $1166.82 \pm 38.88$
& $26.93 \pm 0.71$
& $1759.91 \pm 95.60$
& $31.39 \pm 0.75$
& $2437.39 \pm 131.64$ \\

& \multirow{2}{*}{FT-64}
& Recent 18
& $19.74 \pm 0.44$
& $929.47 \pm 41.18$
& $23.04 \pm 0.60$
& $1311.65 \pm 76.47$
& $26.52 \pm 0.72$
& $\underline{1740.50} \pm 98.80$ \\

&
& Recent + prefix
& $\underline{19.61} \pm 0.25$
& $\underline{877.52} \pm 21.51$
& $\underline{22.72} \pm 0.22$
& $\underline{1215.10} \pm 18.20$
& $\underline{26.09} \pm 0.12$
& $\mathbf{1640.43} \pm 7.27$ \\*

\midrule

% =========================================================
% PEMS04
% =========================================================
\multicolumn{9}{l}{\textbf{PEMS04}} \\*

\textbf{GAS-PFN}
& --
& Mixed
& $\mathbf{22.52} \pm 0.28$
& $\mathbf{1243.43} \pm 13.35$
& $\mathbf{27.16} \pm 0.41$
& $\mathbf{1782.98} \pm 26.45$
& $\mathbf{32.32} \pm 0.53$
& $\underline{2517.88} \pm 54.69$ \\*

\addlinespace[1pt]

\multirow{4}{*}{TimePFN}
& \multirow{2}{*}{FT-32}
& Recent 18
& $27.00 \pm 1.28$
& $1538.41 \pm 83.40$
& $31.26 \pm 0.81$
& $2085.04 \pm 63.11$
& $35.92 \pm 0.52$
& $2720.01 \pm 86.17$ \\

&
& Recent + prefix
& $27.57 \pm 0.44$
& $1712.73 \pm 38.58$
& $32.81 \pm 0.63$
& $2454.12 \pm 74.49$
& $38.33 \pm 0.74$
& $3404.70 \pm 108.27$ \\

& \multirow{2}{*}{FT-64}
& Recent 18
& $\underline{25.04} \pm 0.57$
& $\underline{1389.74} \pm 52.50$
& $\underline{28.68} \pm 0.68$
& $\underline{1828.83} \pm 88.27$
& $\underline{32.74} \pm 0.72$
& $\mathbf{2343.80} \pm 110.14$ \\

&
& Recent + prefix
& $25.57 \pm 0.34$
& $1528.04 \pm 48.25$
& $30.39 \pm 0.52$
& $2186.02 \pm 86.02$
& $35.64 \pm 0.61$
& $3044.37 \pm 118.65$ \\*

\midrule

% =========================================================
% PEMS08
% =========================================================
\multicolumn{9}{l}{\textbf{PEMS08}} \\*

\textbf{GAS-PFN}
& --
& Mixed
& $\mathbf{17.78} \pm 0.19$
& $\mathbf{767.87} \pm 17.31$
& $\mathbf{21.35} \pm 0.36$
& $\mathbf{1112.37} \pm 41.18$
& $\mathbf{25.21} \pm 0.47$
& $\underline{1548.52} \pm 71.32$ \\*

\addlinespace[1pt]

\multirow{4}{*}{TimePFN}
& \multirow{2}{*}{FT-32}
& Recent 18
& $21.71 \pm 1.09$
& $1013.32 \pm 57.76$
& $25.56 \pm 0.63$
& $1418.37 \pm 31.71$
& $29.79 \pm 0.37$
& $1892.39 \pm 56.39$ \\

&
& Recent + prefix
& $21.54 \pm 0.70$
& $998.85 \pm 51.57$
& $24.61 \pm 0.85$
& $1326.38 \pm 78.51$
& $28.43 \pm 1.15$
& $1768.08 \pm 119.92$ \\

& \multirow{2}{*}{FT-64}
& Recent 18
& $20.43 \pm 0.59$
& $914.07 \pm 49.24$
& $23.86 \pm 0.66$
& $1269.17 \pm 71.42$
& $27.72 \pm 0.61$
& $1708.48 \pm 79.46$ \\

&
& Recent + prefix
& $\underline{19.60} \pm 0.12$
& $\underline{864.92} \pm 9.32$
& $\underline{22.39} \pm 0.14$
& $\underline{1151.18} \pm 19.03$
& $\underline{25.69} \pm 0.26$
& $\mathbf{1532.11} \pm 45.59$ \\*

\bottomrule
\end{tabular}
\endgroup
\end{table*}

\subsection{Full SCPT Comparison}

Table~\ref{tab:scpt_road_fraction} expands the main-paper cross-road comparison by reporting complete MAE and MSE results at the cumulative 30-, 60-, and 90-minute horizons. SCPT12 is trained separately for each road split using the full historical record from the indicated fraction of roads, with 10\% reserved for validation and the remaining roads held out for testing. GAS-PFN instead performs zero-shot inference without target-specific parameter updates. The 60-minute results reproduce the main-paper finding that GAS-PFN is strongest when supervised road coverage is limited: it outperforms SCPT12 on all six datasets at the 10\% budget and on five of six datasets at both 20\% and 30\%. The additional horizons show the same broad pattern, while SCPT12 generally becomes more competitive as its access to supervised roads increases. Full preprocessing and split details are provided in Appendix~\ref{app:experimental-details}.

\begin{table*}[!t]
\centering

\caption{
Cross-road comparison of zero-shot GAS-PFN and SCPT12 across supervised road fractions and cumulative 30-, 60-, and 90-minute horizons. Results are mean $\pm$ sample standard deviation over paired seeds 42, 43, and 44. Lower is better; bold denotes the lower unrounded mean within each road fraction, horizon, and metric.
}
\label{tab:scpt_road_fraction}

\begingroup
\scriptsize
\setlength{\tabcolsep}{1.2pt}
\renewcommand{\arraystretch}{1.00}

\begin{tabular}{@{}llcccccc@{}}
\toprule
\multirow{2}{*}{\textbf{Train roads}}
& \multirow{2}{*}{\textbf{Model}}
& \multicolumn{2}{c}{\textbf{30 min}}
& \multicolumn{2}{c}{\textbf{60 min}}
& \multicolumn{2}{c}{\textbf{90 min}} \\
\cmidrule(lr){3-4}
\cmidrule(lr){5-6}
\cmidrule(lr){7-8}
&
& \textbf{MAE} & \textbf{MSE}
& \textbf{MAE} & \textbf{MSE}
& \textbf{MAE} & \textbf{MSE} \\
\midrule

% =========================================================
% METR-LA
% =========================================================

\multicolumn{8}{l}{\textbf{METR-LA}} \\*
\addlinespace[1pt]

\multirow{2}{*}{10\%}
& \textbf{GAS-PFN}
& $\mathbf{3.99} \pm 0.08$
& $\mathbf{68.29} \pm 2.57$
& $\mathbf{4.85} \pm 0.10$
& $\mathbf{97.40} \pm 4.41$
& $\mathbf{5.52} \pm 0.13$
& $\mathbf{119.25} \pm 6.98$ \\

& SCPT12
& $4.32 \pm 0.55$
& $76.25 \pm 7.20$
& $5.48 \pm 1.02$
& $119.35 \pm 19.44$
& $6.36 \pm 1.26$
& $154.20 \pm 26.70$ \\

\cmidrule(lr){1-8}

\multirow{2}{*}{20\%}
& \textbf{GAS-PFN}
& $\mathbf{3.99} \pm 0.13$
& $\mathbf{68.20} \pm 3.57$
& $\mathbf{4.84} \pm 0.17$
& $\mathbf{97.04} \pm 6.11$
& $\mathbf{5.50} \pm 0.23$
& $\mathbf{118.63} \pm 9.13$ \\

& SCPT12
& $4.26 \pm 0.44$
& $77.03 \pm 6.41$
& $5.31 \pm 0.49$
& $119.07 \pm 8.59$
& $6.13 \pm 0.57$
& $152.71 \pm 12.12$ \\

\cmidrule(lr){1-8}

\multirow{2}{*}{30\%}
& \textbf{GAS-PFN}
& $\mathbf{3.99} \pm 0.12$
& $\mathbf{68.07} \pm 3.33$
& $\mathbf{4.83} \pm 0.16$
& $\mathbf{96.40} \pm 5.81$
& $\mathbf{5.49} \pm 0.22$
& $\mathbf{117.47} \pm 8.85$ \\

& SCPT12
& $4.00 \pm 0.07$
& $74.70 \pm 3.02$
& $4.95 \pm 0.15$
& $114.32 \pm 8.67$
& $5.70 \pm 0.20$
& $147.04 \pm 12.63$ \\

\cmidrule(lr){1-8}

\multirow{2}{*}{40\%}
& \textbf{GAS-PFN}
& $4.02 \pm 0.15$
& $\mathbf{68.75} \pm 4.46$
& $4.88 \pm 0.20$
& $\mathbf{97.51} \pm 7.31$
& $5.55 \pm 0.26$
& $\mathbf{119.02} \pm 10.34$ \\

& SCPT12
& $\mathbf{3.87} \pm 0.07$
& $70.62 \pm 0.26$
& $\mathbf{4.72} \pm 0.07$
& $104.84 \pm 2.26$
& $\mathbf{5.38} \pm 0.09$
& $132.73 \pm 3.56$ \\

\cmidrule(lr){1-8}

\multirow{2}{*}{50\%}
& \textbf{GAS-PFN}
& $\mathbf{4.00} \pm 0.15$
& $\mathbf{68.33} \pm 4.35$
& $\mathbf{4.85} \pm 0.19$
& $\mathbf{97.19} \pm 7.27$
& $\mathbf{5.52} \pm 0.25$
& $\mathbf{118.96} \pm 10.31$ \\

& SCPT12
& $4.00 \pm 0.36$
& $74.09 \pm 6.36$
& $4.99 \pm 0.59$
& $111.98 \pm 12.78$
& $5.72 \pm 0.74$
& $141.90 \pm 17.20$ \\

\cmidrule(lr){1-8}

\multirow{2}{*}{60\%}
& \textbf{GAS-PFN}
& $4.02 \pm 0.16$
& $\mathbf{67.90} \pm 3.60$
& $4.87 \pm 0.22$
& $\mathbf{96.49} \pm 6.38$
& $5.55 \pm 0.29$
& $\mathbf{118.19} \pm 9.84$ \\

& SCPT12
& $\mathbf{3.78} \pm 0.15$
& $68.86 \pm 2.77$
& $\mathbf{4.63} \pm 0.18$
& $102.33 \pm 2.79$
& $\mathbf{5.30} \pm 0.21$
& $129.56 \pm 1.86$ \\

\cmidrule(lr){1-8}

\multirow{2}{*}{70\%}
& \textbf{GAS-PFN}
& $4.03 \pm 0.11$
& $\mathbf{67.67} \pm 1.46$
& $4.91 \pm 0.09$
& $\mathbf{96.86} \pm 1.30$
& $5.61 \pm 0.10$
& $\mathbf{119.33} \pm 2.54$ \\

& SCPT12
& $\mathbf{3.84} \pm 0.11$
& $71.82 \pm 1.86$
& $\mathbf{4.76} \pm 0.10$
& $108.71 \pm 2.00$
& $\mathbf{5.50} \pm 0.10$
& $138.44 \pm 1.14$ \\

\cmidrule(lr){1-8}

\multirow{2}{*}{80\%}
& \textbf{GAS-PFN}
& $4.01 \pm 0.40$
& $\mathbf{69.34} \pm 6.23$
& $4.92 \pm 0.46$
& $\mathbf{100.36} \pm 8.83$
& $5.64 \pm 0.53$
& $\mathbf{124.44} \pm 12.57$ \\

& SCPT12
& $\mathbf{3.82} \pm 0.21$
& $71.80 \pm 3.32$
& $\mathbf{4.77} \pm 0.16$
& $109.75 \pm 4.14$
& $\mathbf{5.53} \pm 0.14$
& $140.74 \pm 5.04$ \\

\midrule

% =========================================================
% PEMS-BAY
% =========================================================

\multicolumn{8}{l}{\textbf{PEMS-BAY}} \\*
\addlinespace[1pt]

\multirow{2}{*}{10\%}
& \textbf{GAS-PFN}
& $\mathbf{1.72} \pm 0.04$
& $14.78 \pm 0.35$
& $\mathbf{2.20} \pm 0.04$
& $\mathbf{25.12} \pm 0.38$
& $\mathbf{2.58} \pm 0.04$
& $\mathbf{34.33} \pm 0.44$ \\

& SCPT12
& $1.83 \pm 0.21$
& $\mathbf{12.53} \pm 1.28$
& $2.67 \pm 0.88$
& $26.51 \pm 10.90$
& $3.53 \pm 1.76$
& $45.86 \pm 32.26$ \\

\cmidrule(lr){1-8}

\multirow{2}{*}{20\%}
& \textbf{GAS-PFN}
& $1.73 \pm 0.05$
& $14.95 \pm 0.57$
& $2.21 \pm 0.06$
& $25.41 \pm 0.72$
& $2.60 \pm 0.06$
& $34.76 \pm 0.67$ \\

& SCPT12
& $\mathbf{1.63} \pm 0.04$
& $\mathbf{11.48} \pm 0.45$
& $\mathbf{2.04} \pm 0.06$
& $\mathbf{19.60} \pm 0.96$
& $\mathbf{2.34} \pm 0.07$
& $\mathbf{25.95} \pm 1.10$ \\

\cmidrule(lr){1-8}

\multirow{2}{*}{30\%}
& \textbf{GAS-PFN}
& $1.71 \pm 0.05$
& $14.73 \pm 0.40$
& $2.19 \pm 0.06$
& $25.17 \pm 0.51$
& $2.57 \pm 0.06$
& $34.49 \pm 0.49$ \\

& SCPT12
& $\mathbf{1.54} \pm 0.07$
& $\mathbf{10.57} \pm 0.51$
& $\mathbf{1.94} \pm 0.11$
& $\mathbf{17.50} \pm 0.59$
& $\mathbf{2.25} \pm 0.16$
& $\mathbf{23.05} \pm 0.69$ \\

\cmidrule(lr){1-8}

\multirow{2}{*}{40\%}
& \textbf{GAS-PFN}
& $1.71 \pm 0.06$
& $14.66 \pm 0.69$
& $2.19 \pm 0.08$
& $25.03 \pm 1.11$
& $2.57 \pm 0.09$
& $34.31 \pm 1.34$ \\

& SCPT12
& $\mathbf{1.54} \pm 0.02$
& $\mathbf{10.86} \pm 0.50$
& $\mathbf{1.91} \pm 0.03$
& $\mathbf{18.30} \pm 1.48$
& $\mathbf{2.17} \pm 0.05$
& $\mathbf{24.16} \pm 2.21$ \\

\cmidrule(lr){1-8}

\multirow{2}{*}{50\%}
& \textbf{GAS-PFN}
& $1.68 \pm 0.05$
& $14.09 \pm 0.78$
& $2.15 \pm 0.07$
& $24.01 \pm 1.37$
& $2.52 \pm 0.09$
& $32.87 \pm 1.93$ \\

& SCPT12
& $\mathbf{1.45} \pm 0.01$
& $\mathbf{9.66} \pm 0.44$
& $\mathbf{1.79} \pm 0.01$
& $\mathbf{15.85} \pm 0.43$
& $\mathbf{2.04} \pm 0.01$
& $\mathbf{20.60} \pm 0.25$ \\

\cmidrule(lr){1-8}

\multirow{2}{*}{60\%}
& \textbf{GAS-PFN}
& $1.69 \pm 0.08$
& $14.32 \pm 1.17$
& $2.16 \pm 0.10$
& $24.39 \pm 1.89$
& $2.53 \pm 0.12$
& $33.37 \pm 2.57$ \\

& SCPT12
& $\mathbf{1.45} \pm 0.07$
& $\mathbf{10.00} \pm 0.87$
& $\mathbf{1.81} \pm 0.08$
& $\mathbf{16.86} \pm 1.04$
& $\mathbf{2.07} \pm 0.10$
& $\mathbf{22.77} \pm 1.30$ \\

\cmidrule(lr){1-8}

\multirow{2}{*}{70\%}
& \textbf{GAS-PFN}
& $1.68 \pm 0.08$
& $14.12 \pm 1.31$
& $2.15 \pm 0.10$
& $24.20 \pm 2.12$
& $2.53 \pm 0.11$
& $33.22 \pm 2.80$ \\

& SCPT12
& $\mathbf{1.48} \pm 0.06$
& $\mathbf{10.19} \pm 0.77$
& $\mathbf{1.90} \pm 0.05$
& $\mathbf{17.48} \pm 1.06$
& $\mathbf{2.25} \pm 0.09$
& $\mathbf{23.62} \pm 1.13$ \\

\cmidrule(lr){1-8}

\multirow{2}{*}{80\%}
& \textbf{GAS-PFN}
& $1.72 \pm 0.13$
& $14.48 \pm 1.11$
& $2.20 \pm 0.14$
& $24.58 \pm 1.82$
& $2.58 \pm 0.16$
& $33.57 \pm 2.81$ \\

& SCPT12
& $\mathbf{1.54} \pm 0.11$
& $\mathbf{11.16} \pm 0.78$
& $\mathbf{1.98} \pm 0.09$
& $\mathbf{19.87} \pm 0.73$
& $\mathbf{2.31} \pm 0.09$
& $\mathbf{27.61} \pm 2.25$ \\

\bottomrule
\end{tabular}
\endgroup
\end{table*}

% =====================================================================
% CONTINUED TABLE
% =====================================================================

\begin{table*}[!t]
\centering

{\scriptsize\itshape
Table~\ref{tab:scpt_road_fraction} continued
\par}
\vspace{2pt}

\begingroup
\scriptsize
\setlength{\tabcolsep}{1.2pt}
\renewcommand{\arraystretch}{1.00}

\begin{tabular}{@{}llcccccc@{}}
\toprule
\multirow{2}{*}{\textbf{Train roads}}
& \multirow{2}{*}{\textbf{Model}}
& \multicolumn{2}{c}{\textbf{30 min}}
& \multicolumn{2}{c}{\textbf{60 min}}
& \multicolumn{2}{c}{\textbf{90 min}} \\
\cmidrule(lr){3-4}
\cmidrule(lr){5-6}
\cmidrule(lr){7-8}
&
& \textbf{MAE} & \textbf{MSE}
& \textbf{MAE} & \textbf{MSE}
& \textbf{MAE} & \textbf{MSE} \\
\midrule

% =========================================================
% PEMS03
% =========================================================

\multicolumn{8}{l}{\textbf{PEMS03}} \\*
\addlinespace[1pt]

\multirow{2}{*}{10\%}
& \textbf{GAS-PFN}
& $\mathbf{17.64} \pm 0.69$
& $\mathbf{753.83} \pm 49.48$
& $\mathbf{21.58} \pm 1.02$
& $\mathbf{1167.46} \pm 95.74$
& $\mathbf{26.05} \pm 1.46$
& $\mathbf{1751.57} \pm 173.62$ \\

& SCPT12
& $23.76 \pm 2.41$
& $1474.14 \pm 269.95$
& $30.43 \pm 4.14$
& $2443.89 \pm 623.04$
& $36.68 \pm 5.58$
& $3559.57 \pm 1016.24$ \\

\cmidrule(lr){1-8}

\multirow{2}{*}{20\%}
& \textbf{GAS-PFN}
& $\mathbf{17.80} \pm 0.61$
& $\mathbf{762.80} \pm 46.22$
& $\mathbf{21.77} \pm 0.90$
& $\mathbf{1181.56} \pm 91.13$
& $\mathbf{26.27} \pm 1.31$
& $\mathbf{1771.99} \pm 165.93$ \\

& SCPT12
& $22.57 \pm 2.96$
& $1305.09 \pm 301.00$
& $28.30 \pm 4.85$
& $2095.35 \pm 735.58$
& $34.13 \pm 7.00$
& $3121.79 \pm 1310.33$ \\

\cmidrule(lr){1-8}

\multirow{2}{*}{30\%}
& \textbf{GAS-PFN}
& $\mathbf{17.77} \pm 0.35$
& $\mathbf{757.35} \pm 34.66$
& $\mathbf{21.72} \pm 0.55$
& $\mathbf{1171.21} \pm 74.17$
& $\mathbf{26.20} \pm 0.85$
& $\mathbf{1754.35} \pm 137.57$ \\

& SCPT12
& $19.07 \pm 1.69$
& $952.62 \pm 58.43$
& $23.37 \pm 3.05$
& $1410.53 \pm 130.50$
& $27.82 \pm 5.05$
& $1994.66 \pm 355.93$ \\

\cmidrule(lr){1-8}

\multirow{2}{*}{40\%}
& \textbf{GAS-PFN}
& $17.62 \pm 0.46$
& $\mathbf{747.38} \pm 42.85$
& $21.55 \pm 0.64$
& $1155.99 \pm 85.87$
& $26.00 \pm 0.92$
& $1732.82 \pm 152.98$ \\

& SCPT12
& $\mathbf{16.92} \pm 0.96$
& $798.44 \pm 194.74$
& $\mathbf{19.92} \pm 1.39$
& $\mathbf{1084.35} \pm 268.59$
& $\mathbf{23.00} \pm 1.45$
& $\mathbf{1417.65} \pm 301.37$ \\

\cmidrule(lr){1-8}

\multirow{2}{*}{50\%}
& \textbf{GAS-PFN}
& $17.67 \pm 0.43$
& $\mathbf{748.69} \pm 47.63$
& $21.62 \pm 0.57$
& $1160.46 \pm 95.33$
& $26.08 \pm 0.78$
& $1740.03 \pm 167.95$ \\

& SCPT12
& $\mathbf{16.09} \pm 0.36$
& $748.79 \pm 167.27$
& $\mathbf{18.90} \pm 0.53$
& $\mathbf{999.71} \pm 140.97$
& $\mathbf{21.82} \pm 1.19$
& $\mathbf{1301.00} \pm 73.33$ \\

\cmidrule(lr){1-8}

\multirow{2}{*}{60\%}
& \textbf{GAS-PFN}
& $17.75 \pm 0.80$
& $\mathbf{751.35} \pm 72.78$
& $21.73 \pm 1.01$
& $\mathbf{1166.89} \pm 133.32$
& $26.24 \pm 1.31$
& $1752.80 \pm 226.64$ \\

& SCPT12
& $\mathbf{16.73} \pm 0.71$
& $895.80 \pm 272.09$
& $\mathbf{19.61} \pm 1.16$
& $1176.12 \pm 350.13$
& $\mathbf{22.59} \pm 1.53$
& $\mathbf{1499.05} \pm 426.91$ \\

\cmidrule(lr){1-8}

\multirow{2}{*}{70\%}
& \textbf{GAS-PFN}
& $17.79 \pm 0.17$
& $\mathbf{753.31} \pm 15.29$
& $21.77 \pm 0.23$
& $1168.44 \pm 38.69$
& $26.28 \pm 0.35$
& $1752.22 \pm 76.11$ \\

& SCPT12
& $\mathbf{16.04} \pm 0.27$
& $767.52 \pm 324.62$
& $\mathbf{18.37} \pm 0.45$
& $\mathbf{997.21} \pm 394.86$
& $\mathbf{20.79} \pm 0.67$
& $\mathbf{1261.68} \pm 452.70$ \\

\cmidrule(lr){1-8}

\multirow{2}{*}{80\%}
& \textbf{GAS-PFN}
& $17.02 \pm 0.27$
& $673.95 \pm 43.09$
& $20.86 \pm 0.37$
& $1054.08 \pm 102.41$
& $25.24 \pm 0.55$
& $1595.88 \pm 203.21$ \\

& SCPT12
& $\mathbf{16.24} \pm 0.61$
& $\mathbf{585.95} \pm 3.48$
& $\mathbf{19.12} \pm 0.64$
& $\mathbf{835.66} \pm 23.90$
& $\mathbf{21.94} \pm 0.78$
& $\mathbf{1120.94} \pm 46.64$ \\

\midrule

% =========================================================
% PEMS04
% =========================================================

\multicolumn{8}{l}{\textbf{PEMS04}} \\*
\addlinespace[1pt]

\multirow{2}{*}{10\%}
& \textbf{GAS-PFN}
& $\mathbf{22.53} \pm 0.52$
& $\mathbf{1250.46} \pm 30.00$
& $\mathbf{27.15} \pm 0.73$
& $\mathbf{1787.00} \pm 50.78$
& $\mathbf{32.28} \pm 0.92$
& $\mathbf{2516.84} \pm 88.98$ \\

& SCPT12
& $36.13 \pm 15.56$
& $3029.26 \pm 2380.93$
& $45.05 \pm 22.39$
& $4600.27 \pm 4015.58$
& $53.35 \pm 28.57$
& $6399.14 \pm 5932.10$ \\

\cmidrule(lr){1-8}

\multirow{2}{*}{20\%}
& \textbf{GAS-PFN}
& $\mathbf{22.75} \pm 0.57$
& $\mathbf{1268.18} \pm 40.49$
& $\mathbf{27.39} \pm 0.75$
& $\mathbf{1813.71} \pm 49.37$
& $\mathbf{32.56} \pm 0.89$
& $\mathbf{2557.23} \pm 67.91$ \\

& SCPT12
& $29.56 \pm 3.96$
& $1809.83 \pm 327.14$
& $35.69 \pm 4.77$
& $2581.80 \pm 343.00$
& $42.21 \pm 5.71$
& $3666.12 \pm 451.95$ \\

\cmidrule(lr){1-8}

\multirow{2}{*}{30\%}
& \textbf{GAS-PFN}
& $\mathbf{22.67} \pm 0.25$
& $\mathbf{1259.84} \pm 4.94$
& $\mathbf{27.30} \pm 0.45$
& $\mathbf{1799.57} \pm 15.85$
& $\mathbf{32.46} \pm 0.67$
& $\mathbf{2536.79} \pm 62.14$ \\

& SCPT12
& $25.75 \pm 1.37$
& $1479.77 \pm 47.21$
& $32.39 \pm 2.62$
& $2278.99 \pm 148.44$
& $39.02 \pm 3.91$
& $3326.64 \pm 335.14$ \\

\cmidrule(lr){1-8}

\multirow{2}{*}{40\%}
& \textbf{GAS-PFN}
& $22.59 \pm 0.38$
& $1245.65 \pm 25.52$
& $27.20 \pm 0.59$
& $1782.38 \pm 52.47$
& $32.35 \pm 0.87$
& $2518.22 \pm 114.79$ \\

& SCPT12
& $\mathbf{22.59} \pm 0.29$
& $\mathbf{1191.91} \pm 38.32$
& $\mathbf{26.27} \pm 0.66$
& $\mathbf{1570.22} \pm 72.18$
& $\mathbf{30.00} \pm 1.39$
& $\mathbf{2024.34} \pm 121.34$ \\

\cmidrule(lr){1-8}

\multirow{2}{*}{50\%}
& \textbf{GAS-PFN}
& $22.20 \pm 0.33$
& $1198.10 \pm 18.90$
& $26.80 \pm 0.52$
& $1729.92 \pm 50.51$
& $31.95 \pm 0.78$
& $2459.15 \pm 115.93$ \\

& SCPT12
& $\mathbf{21.69} \pm 0.72$
& $\mathbf{1109.04} \pm 58.69$
& $\mathbf{25.33} \pm 1.18$
& $\mathbf{1482.22} \pm 125.65$
& $\mathbf{29.23} \pm 1.61$
& $\mathbf{1949.34} \pm 186.84$ \\

\cmidrule(lr){1-8}

\multirow{2}{*}{60\%}
& \textbf{GAS-PFN}
& $22.79 \pm 0.29$
& $1258.91 \pm 31.33$
& $27.47 \pm 0.43$
& $1810.84 \pm 59.39$
& $32.69 \pm 0.62$
& $2562.43 \pm 109.23$ \\

& SCPT12
& $\mathbf{22.36} \pm 0.15$
& $\mathbf{1145.27} \pm 15.29$
& $\mathbf{25.64} \pm 0.11$
& $\mathbf{1482.25} \pm 40.99$
& $\mathbf{28.96} \pm 0.27$
& $\mathbf{1889.31} \pm 99.03$ \\

\cmidrule(lr){1-8}

\multirow{2}{*}{70\%}
& \textbf{GAS-PFN}
& $22.88 \pm 0.91$
& $1273.32 \pm 153.18$
& $27.53 \pm 0.95$
& $1828.01 \pm 199.31$
& $32.70 \pm 1.14$
& $2581.52 \pm 265.26$ \\

& SCPT12
& $\mathbf{22.69} \pm 1.13$
& $\mathbf{1197.25} \pm 138.11$
& $\mathbf{26.18} \pm 1.38$
& $\mathbf{1581.78} \pm 199.02$
& $\mathbf{29.77} \pm 1.78$
& $\mathbf{2056.09} \pm 306.52$ \\

\cmidrule(lr){1-8}

\multirow{2}{*}{80\%}
& \textbf{GAS-PFN}
& $\mathbf{22.02} \pm 0.55$
& $\mathbf{1170.51} \pm 53.70$
& $\mathbf{26.58} \pm 0.52$
& $\mathbf{1703.73} \pm 73.83$
& $\mathbf{31.70} \pm 0.61$
& $\mathbf{2430.69} \pm 144.66$ \\

& SCPT12
& $22.97 \pm 2.82$
& $1193.95 \pm 169.60$
& $28.14 \pm 5.41$
& $1778.84 \pm 479.19$
& $32.90 \pm 7.47$
& $2447.87 \pm 803.34$ \\

\bottomrule
\end{tabular}
\endgroup
\end{table*}

% =====================================================================
% CONTINUED TABLE
% =====================================================================

\begin{table*}[!t]
\centering

{\scriptsize\itshape
Table~\ref{tab:scpt_road_fraction} continued
\par}
\vspace{2pt}

\begingroup
\scriptsize
\setlength{\tabcolsep}{1.2pt}
\renewcommand{\arraystretch}{1.00}

\begin{tabular}{@{}llcccccc@{}}
\toprule
\multirow{2}{*}{\textbf{Train roads}}
& \multirow{2}{*}{\textbf{Model}}
& \multicolumn{2}{c}{\textbf{30 min}}
& \multicolumn{2}{c}{\textbf{60 min}}
& \multicolumn{2}{c}{\textbf{90 min}} \\
\cmidrule(lr){3-4}
\cmidrule(lr){5-6}
\cmidrule(lr){7-8}
&
& \textbf{MAE} & \textbf{MSE}
& \textbf{MAE} & \textbf{MSE}
& \textbf{MAE} & \textbf{MSE} \\
\midrule

% =========================================================
% PEMS07
% =========================================================

\multicolumn{8}{l}{\textbf{PEMS07}} \\*
\addlinespace[1pt]

\multirow{2}{*}{10\%}
& \textbf{GAS-PFN}
& $\mathbf{25.21} \pm 0.57$
& $\mathbf{1502.58} \pm 44.38$
& $\mathbf{30.89} \pm 0.80$
& $\mathbf{2238.09} \pm 78.90$
& $\mathbf{37.23} \pm 1.05$
& $\mathbf{3257.79} \pm 133.56$ \\

& SCPT12
& $28.34 \pm 0.99$
& $1796.27 \pm 117.07$
& $35.22 \pm 0.83$
& $2763.04 \pm 254.87$
& $42.19 \pm 2.30$
& $3988.32 \pm 571.39$ \\

\cmidrule(lr){1-8}

\multirow{2}{*}{20\%}
& \textbf{GAS-PFN}
& $\mathbf{25.36} \pm 0.65$
& $\mathbf{1522.72} \pm 51.96$
& $\mathbf{31.05} \pm 0.90$
& $\mathbf{2263.22} \pm 94.49$
& $\mathbf{37.41} \pm 1.16$
& $\mathbf{3288.00} \pm 156.24$ \\

& SCPT12
& $27.62 \pm 2.68$
& $1711.78 \pm 330.57$
& $35.67 \pm 7.02$
& $2854.59 \pm 1022.40$
& $43.91 \pm 11.83$
& $4411.63 \pm 2085.96$ \\

\cmidrule(lr){1-8}

\multirow{2}{*}{30\%}
& \textbf{GAS-PFN}
& $\mathbf{25.32} \pm 0.56$
& $\mathbf{1526.94} \pm 39.60$
& $\mathbf{31.01} \pm 0.79$
& $\mathbf{2269.69} \pm 80.99$
& $\mathbf{37.37} \pm 1.03$
& $\mathbf{3296.82} \pm 142.17$ \\

& SCPT12
& $28.18 \pm 7.70$
& $1839.26 \pm 877.65$
& $36.26 \pm 14.41$
& $3079.99 \pm 2201.38$
& $43.97 \pm 20.70$
& $4604.30 \pm 3925.94$ \\

\cmidrule(lr){1-8}

\multirow{2}{*}{40\%}
& \textbf{GAS-PFN}
& $25.34 \pm 0.49$
& $1531.86 \pm 44.88$
& $31.01 \pm 0.69$
& $2273.58 \pm 92.14$
& $37.36 \pm 0.88$
& $3299.81 \pm 156.95$ \\

& SCPT12
& $\mathbf{23.31} \pm 0.38$
& $\mathbf{1328.37} \pm 88.44$
& $\mathbf{26.49} \pm 0.63$
& $\mathbf{1681.33} \pm 123.01$
& $\mathbf{29.68} \pm 0.73$
& $\mathbf{2080.28} \pm 131.83$ \\

\cmidrule(lr){1-8}

\multirow{2}{*}{50\%}
& \textbf{GAS-PFN}
& $25.56 \pm 0.43$
& $1563.00 \pm 34.29$
& $31.25 \pm 0.55$
& $2315.38 \pm 57.90$
& $37.60 \pm 0.69$
& $3350.11 \pm 101.54$ \\

& SCPT12
& $\mathbf{24.90} \pm 1.76$
& $\mathbf{1373.35} \pm 109.85$
& $\mathbf{28.91} \pm 2.25$
& $\mathbf{1814.06} \pm 196.47$
& $\mathbf{33.06} \pm 2.74$
& $\mathbf{2345.67} \pm 331.25$ \\

\cmidrule(lr){1-8}

\multirow{2}{*}{60\%}
& \textbf{GAS-PFN}
& $25.34 \pm 0.76$
& $1541.25 \pm 85.62$
& $31.01 \pm 0.94$
& $2285.99 \pm 116.30$
& $37.33 \pm 1.08$
& $3305.68 \pm 149.14$ \\

& SCPT12
& $\mathbf{22.97} \pm 1.18$
& $\mathbf{1256.87} \pm 103.71$
& $\mathbf{26.27} \pm 1.76$
& $\mathbf{1622.91} \pm 186.12$
& $\mathbf{29.72} \pm 2.71$
& $\mathbf{2058.34} \pm 338.67$ \\

\cmidrule(lr){1-8}

\multirow{2}{*}{70\%}
& \textbf{GAS-PFN}
& $25.02 \pm 1.08$
& $1499.77 \pm 146.00$
& $30.70 \pm 1.35$
& $2241.97 \pm 220.56$
& $37.03 \pm 1.56$
& $3262.10 \pm 295.05$ \\

& SCPT12
& $\mathbf{22.13} \pm 1.08$
& $\mathbf{1185.12} \pm 150.90$
& $\mathbf{25.42} \pm 1.59$
& $\mathbf{1525.59} \pm 236.48$
& $\mathbf{28.66} \pm 2.24$
& $\mathbf{1899.34} \pm 347.03$ \\

\cmidrule(lr){1-8}

\multirow{2}{*}{80\%}
& \textbf{GAS-PFN}
& $25.04 \pm 0.45$
& $1470.91 \pm 77.77$
& $30.71 \pm 0.61$
& $2198.89 \pm 100.53$
& $37.08 \pm 0.76$
& $3212.95 \pm 129.80$ \\

& SCPT12
& $\mathbf{22.85} \pm 0.57$
& $\mathbf{1220.04} \pm 76.38$
& $\mathbf{26.58} \pm 1.11$
& $\mathbf{1627.09} \pm 117.81$
& $\mathbf{30.16} \pm 1.66$
& $\mathbf{2066.06} \pm 188.88$ \\

\midrule

% =========================================================
% PEMS08
% =========================================================

\multicolumn{8}{l}{\textbf{PEMS08}} \\*
\addlinespace[1pt]

\multirow{2}{*}{10\%}
& \textbf{GAS-PFN}
& $\mathbf{17.87} \pm 0.47$
& $\mathbf{776.64} \pm 49.05$
& $\mathbf{21.44} \pm 0.53$
& $\mathbf{1124.74} \pm 73.40$
& $\mathbf{25.30} \pm 0.58$
& $\mathbf{1564.47} \pm 107.52$ \\

& SCPT12
& $28.06 \pm 4.89$
& $1559.90 \pm 508.94$
& $34.78 \pm 6.31$
& $2366.32 \pm 727.39$
& $42.60 \pm 8.65$
& $3647.52 \pm 1272.74$ \\

\cmidrule(lr){1-8}

\multirow{2}{*}{20\%}
& \textbf{GAS-PFN}
& $\mathbf{17.66} \pm 0.59$
& $\mathbf{761.56} \pm 58.90$
& $\mathbf{21.19} \pm 0.68$
& $\mathbf{1101.24} \pm 90.92$
& $\mathbf{25.02} \pm 0.79$
& $\mathbf{1530.77} \pm 135.73$ \\

& SCPT12
& $23.11 \pm 1.36$
& $1123.62 \pm 119.80$
& $28.79 \pm 2.68$
& $1749.58 \pm 228.90$
& $34.24 \pm 3.86$
& $2518.53 \pm 380.07$ \\

\cmidrule(lr){1-8}

\multirow{2}{*}{30\%}
& \textbf{GAS-PFN}
& $\mathbf{17.72} \pm 0.50$
& $\mathbf{768.85} \pm 48.42$
& $\mathbf{21.25} \pm 0.55$
& $\mathbf{1110.83} \pm 76.49$
& $\mathbf{25.07} \pm 0.63$
& $\mathbf{1541.61} \pm 116.88$ \\

& SCPT12
& $22.51 \pm 4.23$
& $1105.21 \pm 347.21$
& $27.05 \pm 4.53$
& $1599.94 \pm 455.89$
& $31.53 \pm 4.44$
& $2171.71 \pm 513.66$ \\

\cmidrule(lr){1-8}

\multirow{2}{*}{40\%}
& \textbf{GAS-PFN}
& $\mathbf{17.80} \pm 0.64$
& $\mathbf{783.97} \pm 62.80$
& $\mathbf{21.34} \pm 0.72$
& $\mathbf{1129.80} \pm 96.65$
& $\mathbf{25.17} \pm 0.85$
& $\mathbf{1564.52} \pm 148.20$ \\

& SCPT12
& $19.53 \pm 2.07$
& $849.91 \pm 149.35$
& $23.75 \pm 3.85$
& $1255.96 \pm 342.69$
& $27.85 \pm 5.35$
& $1739.50 \pm 577.39$ \\

\cmidrule(lr){1-8}

\multirow{2}{*}{50\%}
& \textbf{GAS-PFN}
& $\mathbf{18.12} \pm 1.09$
& $\mathbf{815.00} \pm 83.95$
& $\mathbf{21.71} \pm 1.25$
& $\mathbf{1171.62} \pm 124.00$
& $\mathbf{25.60} \pm 1.49$
& $\mathbf{1621.95} \pm 190.02$ \\

& SCPT12
& $19.80 \pm 0.48$
& $908.42 \pm 84.29$
& $24.79 \pm 3.82$
& $1445.54 \pm 461.72$
& $29.59 \pm 6.46$
& $2086.29 \pm 847.38$ \\

\cmidrule(lr){1-8}

\multirow{2}{*}{60\%}
& \textbf{GAS-PFN}
& $\mathbf{18.00} \pm 1.43$
& $809.61 \pm 96.05$
& $21.50 \pm 1.51$
& $1154.59 \pm 126.17$
& $25.31 \pm 1.65$
& $1591.03 \pm 178.20$ \\

& SCPT12
& $18.44 \pm 0.89$
& $\mathbf{777.06} \pm 52.34$
& $\mathbf{21.32} \pm 0.60$
& $\mathbf{1038.83} \pm 15.46$
& $\mathbf{24.29} \pm 0.57$
& $\mathbf{1345.00} \pm 24.03$ \\

\cmidrule(lr){1-8}

\multirow{2}{*}{70\%}
& \textbf{GAS-PFN}
& $\mathbf{17.98} \pm 1.71$
& $\mathbf{780.12} \pm 142.08$
& $\mathbf{21.38} \pm 1.99$
& $\mathbf{1097.55} \pm 204.06$
& $\mathbf{25.14} \pm 2.29$
& $\mathbf{1509.89} \pm 284.98$ \\

& SCPT12
& $18.46 \pm 1.33$
& $813.42 \pm 82.73$
& $22.15 \pm 0.82$
& $1148.50 \pm 74.68$
& $26.17 \pm 1.28$
& $1577.58 \pm 200.51$ \\

\cmidrule(lr){1-8}

\multirow{2}{*}{80\%}
& \textbf{GAS-PFN}
& $\mathbf{18.47} \pm 3.02$
& $\mathbf{804.82} \pm 261.39$
& $\mathbf{21.99} \pm 3.66$
& $\mathbf{1140.90} \pm 382.05$
& $\mathbf{25.84} \pm 4.35$
& $\mathbf{1580.26} \pm 541.06$ \\

& SCPT12
& $20.90 \pm 3.46$
& $943.24 \pm 281.56$
& $26.31 \pm 4.25$
& $1508.69 \pm 401.19$
& $31.85 \pm 4.83$
& $2248.25 \pm 552.31$ \\

\bottomrule
\end{tabular}
\endgroup
\end{table*}

\section{Ablation Tests}
\label{app:ablations}
Table~\ref{tab:frozen_ablation} presents targeted frozen-checkpoint ablations of three components: dual-branch fusion, mixed-context support, and hybrid neighborhood routing. Across the resulting 36 dataset--horizon--metric settings, the full learned mixture achieves the best mean in 23 cases and ranks among the top two in 35. It outperforms both individual experts in 34 settings, while either learned or fixed equal fusion is best in 32, supporting the use of complementary self-channel and multi-channel representations. Removing the historical supports degrades performance in all 36 settings, indicating that the trained model makes effective use of its mixed-context memory. The hybrid router also outperforms graph-only routing in 33 settings, although these gains are generally smaller and remain dataset-dependent. Because these variants use frozen checkpoints, they isolate the
inference-time contribution of each component.

\begin{table*}[!t]
\centering

\caption{
Frozen-checkpoint ablations of branch fusion, context construction, and neighborhood routing across six datasets and cumulative 30-, 60-, and 90-minute horizons. Results are mean $\pm$ sample standard deviation over seeds 42, 43, and 44. Lower is better; bold and underlined values denote the best and second-best unrounded means, respectively.
}
\label{tab:frozen_ablation}

\begingroup
\scriptsize
\setlength{\tabcolsep}{0.9pt}
\renewcommand{\arraystretch}{1.03}

\begin{tabular}{@{}lcccccc@{}}
\toprule
\multirow{2}{*}{\textbf{Variant}}
& \multicolumn{2}{c}{\textbf{30 min}}
& \multicolumn{2}{c}{\textbf{60 min}}
& \multicolumn{2}{c}{\textbf{90 min}} \\
\cmidrule(lr){2-3}
\cmidrule(lr){4-5}
\cmidrule(lr){6-7}
& \textbf{MAE} & \textbf{MSE}
& \textbf{MAE} & \textbf{MSE}
& \textbf{MAE} & \textbf{MSE} \\
\midrule

% =========================================================
% METR-LA
% =========================================================
\multicolumn{7}{l}{\textbf{METR-LA}} \\*

\textbf{Learned mixture (full model)}
& $\mathbf{3.97} \pm 0.07$
& $\underline{68.31} \pm 2.35$
& $\mathbf{4.83} \pm 0.07$
& $\underline{97.61} \pm 3.58$
& $\mathbf{5.50} \pm 0.06$
& $119.69 \pm 5.08$ \\

Self expert only
& $4.09 \pm 0.03$
& $71.71 \pm 1.58$
& $4.97 \pm 0.03$
& $103.28 \pm 2.50$
& $5.66 \pm 0.04$
& $128.33 \pm 3.91$ \\

Multi expert only
& $4.03 \pm 0.10$
& $69.13 \pm 2.85$
& $4.89 \pm 0.11$
& $98.14 \pm 4.12$
& $5.56 \pm 0.10$
& $\underline{119.46} \pm 5.42$ \\

Fixed equal fusion
& $3.98 \pm 0.06$
& $68.39 \pm 1.87$
& $4.84 \pm 0.05$
& $97.89 \pm 2.88$
& $5.51 \pm 0.05$
& $120.52 \pm 4.47$ \\

\addlinespace[1pt]

Recent context only
& $4.09 \pm 0.08$
& $72.17 \pm 2.67$
& $4.97 \pm 0.06$
& $101.94 \pm 3.29$
& $5.64 \pm 0.07$
& $124.20 \pm 5.94$ \\

Graph similarity only
& $\underline{3.98} \pm 0.07$
& $\mathbf{68.20} \pm 2.38$
& $\underline{4.83} \pm 0.07$
& $\mathbf{97.37} \pm 3.43$
& $\underline{5.50} \pm 0.06$
& $\mathbf{119.43} \pm 4.78$ \\*

\midrule

% =========================================================
% PEMS-BAY
% =========================================================
\multicolumn{7}{l}{\textbf{PEMS-BAY}} \\*

\textbf{Learned mixture (full model)}
& $\mathbf{1.73} \pm 0.01$
& $\mathbf{14.85} \pm 0.12$
& $\mathbf{2.21} \pm 0.02$
& $\mathbf{25.22} \pm 0.11$
& $\mathbf{2.59} \pm 0.01$
& $\mathbf{34.45} \pm 0.09$ \\

Self expert only
& $1.80 \pm 0.02$
& $15.95 \pm 0.07$
& $2.29 \pm 0.01$
& $27.11 \pm 0.13$
& $2.67 \pm 0.01$
& $37.00 \pm 0.28$ \\

Multi expert only
& $1.76 \pm 0.02$
& $15.03 \pm 0.18$
& $2.23 \pm 0.03$
& $25.37 \pm 0.22$
& $2.61 \pm 0.02$
& $\underline{34.54} \pm 0.13$ \\

Fixed equal fusion
& $\underline{1.74} \pm 0.01$
& $\underline{14.85} \pm 0.08$
& $2.22 \pm 0.01$
& $\underline{25.31} \pm 0.15$
& $2.60 \pm 0.01$
& $34.63 \pm 0.26$ \\

\addlinespace[1pt]

Recent context only
& $1.92 \pm 0.00$
& $18.07 \pm 0.25$
& $2.43 \pm 0.00$
& $30.05 \pm 0.35$
& $2.85 \pm 0.01$
& $40.90 \pm 0.74$ \\

Graph similarity only
& $\underline{1.74} \pm 0.01$
& $14.93 \pm 0.12$
& $\underline{2.21} \pm 0.02$
& $25.36 \pm 0.16$
& $\underline{2.59} \pm 0.01$
& $34.64 \pm 0.07$ \\*

\midrule

% =========================================================
% PEMS03
% =========================================================
\multicolumn{7}{l}{\textbf{PEMS03}} \\*

\textbf{Learned mixture (full model)}
& $\underline{17.57} \pm 0.60$
& $\underline{749.58} \pm 36.52$
& $\underline{21.51} \pm 0.92$
& $\underline{1160.61} \pm 69.29$
& $\underline{25.97} \pm 1.35$
& $\underline{1740.93} \pm 129.80$ \\

Self expert only
& $18.19 \pm 0.27$
& $794.55 \pm 10.20$
& $22.67 \pm 0.30$
& $1267.97 \pm 21.84$
& $27.57 \pm 0.38$
& $1925.85 \pm 39.88$ \\

Multi expert only
& $17.85 \pm 0.67$
& $770.64 \pm 46.44$
& $21.81 \pm 1.17$
& $1189.97 \pm 100.60$
& $26.29 \pm 1.81$
& $1779.50 \pm 194.49$ \\

Fixed equal fusion
& $\mathbf{17.54} \pm 0.49$
& $\mathbf{745.94} \pm 24.89$
& $\mathbf{21.49} \pm 0.72$
& $\mathbf{1156.30} \pm 44.52$
& $\mathbf{25.93} \pm 1.04$
& $\mathbf{1731.77} \pm 84.19$ \\

\addlinespace[1pt]

Recent context only
& $23.19 \pm 1.36$
& $1231.18 \pm 141.04$
& $29.60 \pm 1.42$
& $2045.32 \pm 186.21$
& $35.33 \pm 1.40$
& $2947.72 \pm 215.51$ \\

Graph similarity only
& $17.72 \pm 0.61$
& $766.13 \pm 38.67$
& $21.79 \pm 0.94$
& $1198.56 \pm 75.07$
& $26.37 \pm 1.38$
& $1807.21 \pm 141.31$ \\*

\midrule

% =========================================================
% PEMS04
% =========================================================
\multicolumn{7}{l}{\textbf{PEMS04}} \\*

\textbf{Learned mixture (full model)}
& $\underline{22.52} \pm 0.28$
& $\mathbf{1243.43} \pm 13.35$
& $\underline{27.16} \pm 0.41$
& $\mathbf{1782.98} \pm 26.45$
& $\underline{32.32} \pm 0.53$
& $\mathbf{2517.88} \pm 54.69$ \\

Self expert only
& $23.00 \pm 0.08$
& $1302.06 \pm 10.46$
& $27.82 \pm 0.13$
& $1890.23 \pm 23.07$
& $33.11 \pm 0.23$
& $2677.98 \pm 42.77$ \\

Multi expert only
& $22.94 \pm 0.28$
& $1277.37 \pm 16.97$
& $27.72 \pm 0.49$
& $1836.38 \pm 45.04$
& $33.03 \pm 0.71$
& $2594.46 \pm 93.05$ \\

Fixed equal fusion
& $\mathbf{22.49} \pm 0.22$
& $\underline{1243.53} \pm 7.52$
& $\mathbf{27.12} \pm 0.30$
& $\underline{1785.02} \pm 15.03$
& $\mathbf{32.27} \pm 0.39$
& $\underline{2520.70} \pm 42.99$ \\

\addlinespace[1pt]

Recent context only
& $25.68 \pm 0.14$
& $1505.82 \pm 14.72$
& $32.36 \pm 0.10$
& $2366.55 \pm 50.17$
& $39.24 \pm 0.39$
& $3515.84 \pm 122.61$ \\

Graph similarity only
& $22.56 \pm 0.26$
& $1248.64 \pm 11.49$
& $27.17 \pm 0.33$
& $1787.35 \pm 15.84$
& $32.33 \pm 0.41$
& $2524.79 \pm 35.37$ \\*

\midrule

% =========================================================
% PEMS07
% =========================================================
\multicolumn{7}{l}{\textbf{PEMS07}} \\*

\textbf{Learned mixture (full model)}
& $\mathbf{25.15} \pm 0.54$
& $\mathbf{1493.64} \pm 33.87$
& $\mathbf{30.81} \pm 0.75$
& $\mathbf{2223.93} \pm 56.13$
& $\mathbf{37.15} \pm 0.98$
& $\mathbf{3239.16} \pm 94.62$ \\

Self expert only
& $26.31 \pm 0.25$
& $1609.67 \pm 23.75$
& $32.94 \pm 0.31$
& $2504.72 \pm 57.46$
& $40.11 \pm 0.35$
& $3734.87 \pm 87.14$ \\

Multi expert only
& $25.44 \pm 0.56$
& $1520.44 \pm 43.87$
& $31.03 \pm 0.90$
& $2245.55 \pm 83.57$
& $\underline{37.29} \pm 1.24$
& $\underline{3246.51} \pm 132.05$ \\

Fixed equal fusion
& $\underline{25.18} \pm 0.43$
& $\underline{1495.28} \pm 21.85$
& $\underline{30.94} \pm 0.57$
& $\underline{2240.81} \pm 42.61$
& $37.36 \pm 0.73$
& $3272.79 \pm 78.99$ \\

\addlinespace[1pt]

Recent context only
& $32.65 \pm 2.04$
& $2302.57 \pm 243.58$
& $42.05 \pm 2.36$
& $3820.24 \pm 365.82$
& $50.86 \pm 2.37$
& $5625.98 \pm 449.48$ \\

Graph similarity only
& $25.39 \pm 0.52$
& $1529.15 \pm 36.90$
& $31.26 \pm 0.75$
& $2298.15 \pm 64.31$
& $37.84 \pm 1.02$
& $3372.69 \pm 109.55$ \\*

\midrule

% =========================================================
% PEMS08
% =========================================================
\multicolumn{7}{l}{\textbf{PEMS08}} \\*

\textbf{Learned mixture (full model)}
& $\mathbf{17.78} \pm 0.19$
& $\mathbf{767.87} \pm 17.31$
& $\mathbf{21.35} \pm 0.36$
& $\mathbf{1112.37} \pm 41.18$
& $\mathbf{25.21} \pm 0.47$
& $\underline{1548.52} \pm 71.32$ \\

Self expert only
& $18.55 \pm 0.11$
& $831.76 \pm 17.25$
& $22.59 \pm 0.10$
& $1243.81 \pm 34.50$
& $26.96 \pm 0.11$
& $1775.85 \pm 54.62$ \\

Multi expert only
& $17.94 \pm 0.45$
& $\underline{773.35} \pm 26.72$
& $21.54 \pm 0.80$
& $\underline{1114.12} \pm 57.73$
& $25.43 \pm 0.97$
& $\mathbf{1543.48} \pm 89.51$ \\

Fixed equal fusion
& $\underline{17.84} \pm 0.14$
& $774.38 \pm 17.10$
& $\underline{21.47} \pm 0.27$
& $1128.82 \pm 39.72$
& $\underline{25.38} \pm 0.34$
& $1577.33 \pm 68.79$ \\

\addlinespace[1pt]

Recent context only
& $20.98 \pm 0.78$
& $1006.39 \pm 59.22$
& $26.20 \pm 1.21$
& $1481.44 \pm 98.07$
& $31.66 \pm 1.46$
& $2099.19 \pm 142.38$ \\

Graph similarity only
& $17.88 \pm 0.19$
& $777.69 \pm 17.73$
& $21.50 \pm 0.35$
& $1130.83 \pm 39.46$
& $25.41 \pm 0.42$
& $1577.23 \pm 65.29$ \\*

\bottomrule
\end{tabular}
\endgroup
\end{table*}

\paragraph{AI-use disclosure.}
Generative AI tools were used for language editing, \LaTeX{} formatting
assistance, and manuscript review. The authors verified and take
responsibility for all content.

\raggedbottom

% Use the same BibTeX database as the main paper.
% aaai2027.sty sets the AAAI bibliography style.
\bibliography{ref}